\documentclass{article}
\PassOptionsToPackage{numbers,square,comma,sort&compress}{natbib}
\usepackage[main, final]{neurips_2026}

\makeatletter
\renewcommand{\@notice}{}
\renewcommand{\@toptitlebar}{}
\renewcommand{\@bottomtitlebar}{}
\makeatother

\usepackage{siunitx}
\usepackage{lipsum}

\usepackage{pgfplots}
\usepackage{pgfplotstable}
\pgfplotsset{compat=1.18}

\usepackage{tabularx}
\usepackage{adjustbox}
\usepackage{physics}
\usepackage{csquotes}
\usepackage{colortbl}
\usepackage[dvipsnames]{xcolor}
\usepackage{pifont}
\usepackage{graphicx}
\usepackage{booktabs}
\usepackage{caption}
\usepackage{amssymb}
\usepackage{hyperref}

\AtEndPreamble{
    \usepackage[capitalize]{cleveref}
    \crefname{section}{Sec.}{Secs.}
    \Crefname{section}{Section}{Sections}
    \Crefname{table}{Table}{Tables}
    \crefname{table}{Tab.}{Tabs.}
}

\newcommand{\ours}{Volt}
\newcommand{\PAR}[1]{\vskip6pt \noindent {\bf #1~}}
\newcommand{\PARbegin}[1]{\noindent {\bf #1~}}

\newcommand{\httpsurl}[1]{\href{https://#1}{\texttt{#1}}}

\definecolor{noreproduce}{gray}{0.6}
\definecolor{mutedblue}{HTML}{0072B2}
\definecolor{mutedred}{HTML}{D55E00}
\definecolor{mutedteal}{HTML}{009E73}
\definecolor{mutedorange}{HTML}{E69F00}
\definecolor{mutedpurple}{HTML}{CC79A7}
\definecolor{mutedgray}{HTML}{4D4D4D}
\definecolor{mutedbrown}{HTML}{8C564B}
\definecolor{accentred}{HTML}{CC0000}
\definecolor{accentblue}{HTML}{0072B2}

\newcolumntype{Y}{>{\centering\arraybackslash}X}
\newcolumntype{s}{>{\centering}p{2pt}}
\newcolumntype{L}{>{\raggedright\arraybackslash}X}

\definecolor{dino}{HTML}{DBEAFE}
\definecolor{dino_border}{HTML}{51A2FF}
\definecolor{backbone}{HTML}{FEF3C6}
\definecolor{backbone_border}{HTML}{FFB900}
\makeatletter
\newcommand*\fsize{\dimexpr\f@size pt\relax}\makeatother
\DeclareRobustCommand{\colorsquare}[2]{\raisebox{-0.075\fsize}{\tikz\path[draw=#1,fill=#2, line width=0.075*\fsize, rounded corners=0.1*\fsize] (0,0) rectangle (0.625*\fsize,0.625*\fsize);}}

\usepackage[normalem]{ulem}
\usepackage{multirow}

\makeatletter
\newcommand\notsotiny{\@setfontsize\notsotiny{6.5}{7.5}}
\makeatother

\setcitestyle{numbers,square,comma,sort&compress}
\title{Volume Transformer: Revisiting Vanilla Transformers for 3D Scene Understanding}

\author{
\centering
\begin{tabular}{c}
\\[1ex]
\textbf{Kadir Yilmaz}$^{1,\star}$ \quad
\textbf{Adrian Kruse}$^{1,\star}$ \quad
\textbf{Tristan Höfer}$^{1}$ \quad
\textbf{Daan de Geus}$^{2}$ \quad
\textbf{Bastian Leibe}$^{1}$ \\[1ex]
{\normalfont $^{1}$RWTH Aachen University \qquad
$^{2}$Eindhoven University of Technology} \\[1ex]
{\normalfont $^{\star}$Equal contribution} \\[1ex]
{\normalfont \httpsurl{vision.rwth-aachen.de/Volt}}
\end{tabular}
}

\begin{document}

\maketitle

\begin{abstract}
Transformers have become a common foundation across deep learning, yet 3D scene understanding still relies on specialized backbones with strong domain priors.
This isolates the field from the broader Transformer ecosystem, limiting the transfer of research advances from other domains and the benefits of increasingly optimized software and hardware stacks.
To bridge this gap, we propose the Volume Transformer (Volt), which adapts the vanilla Transformer encoder to 3D scenes with minimal modifications.
Specifically, Volt partitions 3D scenes into volumetric patch tokens, processes them with full global self-attention, and injects positional information via 3D rotary positional embeddings (RoPE).
Our initial experiments reveal that naively training Volt on standard 3D benchmarks leads to poor generalization, highlighting the limited scale of current 3D supervision.
To overcome this, we introduce a data-efficient training recipe based on strong 3D augmentations, regularization, and distillation from a convolutional teacher, making Volt competitive with state-of-the-art methods.
We then scale supervision through joint training on multiple datasets and show that Volt benefits more from increased scale than domain-specific 3D backbones, achieving state-of-the-art results on several indoor and outdoor semantic segmentation benchmarks.
Finally, as a drop-in backbone in a standard 3D instance segmentation pipeline, Volt also sets a new state of the art, highlighting its potential as a simple, scalable, and general-purpose backbone for 3D scene understanding.
\end{abstract}
\section{Introduction}
The Transformer architecture~\cite{vaswani2017attention} has emerged as a unifying building block in deep learning.
Originally introduced for natural language processing, it has since become the standard architecture across a wide range of domains~\cite{dosovitskiy2021vit,brown2020gpt3,radford2023whisper,radford2021clip}.
This widespread adoption has created a shared research ecosystem, where advancements in one domain often transfer to others with minimal changes~\cite{su2024roformer,he2022mae,henry2020qknorm}.
At the same time, a mature hardware and software stack has emerged: modern accelerators are explicitly optimized for Transformer workloads~\cite{nvidia_hopper_blog}, and efficient implementations such as FlashAttention~\cite{dao2022flashattention,dao2023flashattention2,shah2024flashattention3} leverage these specialized hardware capabilities to maximize throughput.
Altogether, these properties make Transformers a compelling design choice for building scalable models over specialized, domain-specific architectures.

In contrast to the cross-domain convergence around Transformers, 3D scene understanding remains dominated by specialized architectures.
Fully convolutional 3D U-Nets~\cite{choy20194d,zhu2021cylindrical} have long served as the backbone of choice and continue to be highly competitive, particularly for instance-level tasks~\cite{schult2023mask3d,kolodiazhnyi2024oneformer3d,yilmaz24mask4former}.
More recent works, motivated by the success of the Vision Transformer (ViT)~\cite{dosovitskiy2021vit}, have incorporated attention mechanisms into 3D backbones while retaining the U-Net design and restricting the attention to local windows~\cite{zhao2021ptv1,wu2022ptv2,wu2024ptv3,yang2023swin3d,lai2022stratified,lai2023spherical,yue2025litept}.
For instance, the current state of the art, PTv3~\cite{wu2024ptv3}, utilizes space-filling curves to group points for local attention and uses 3D convolutions for positional encoding and feature aggregation~\cite{yue2025litept}.
As a result, both the earlier convolutional and recent attention-based 3D backbones use domain-specific local operations and depend on multi-resolution architectures to model long-range dependencies.
These design choices provide strong inductive biases that are effective in low-data regimes, but can become a limitation as supervision scales, as demonstrated in 2D vision~\cite{dosovitskiy2021vit}.
Furthermore, the resulting specialized architectures isolate 3D scene understanding from the broader Transformer ecosystem, limiting both the transfer of research advances from other domains and the adoption of Transformer-specific hardware and software optimizations.

To bridge this gap, we revisit the vanilla Transformer architecture and apply it to 3D scene understanding with minimal modifications.
Given an input point cloud, we first impose a grid structure by subsampling it on a 3D voxel grid.
Then, analogous to ViT patch embedding, we partition the voxelized space into non-overlapping cubic patches, flatten each patch, and linearly project it to the Transformer's embedding dimension, yielding a sequence of tokens.
These tokens are processed by a standard Transformer encoder with global self-attention.
To inject positional information, we extend rotary positional embeddings~\cite{su2024roformer} (RoPE), which have proven effective in 1D language~\cite{grattafiori2024llama3,team2025gemma3} and 2D vision~\cite{simeoni2025dinov3,carion2025sam3} data, to 3D volumes.
We call the resulting architecture the \emph{Volume Transformer} (\ours{}) and apply it to 3D segmentation tasks using a lightweight decoder to produce dense predictions.

A common concern is that full global self-attention may be prohibitively expensive for 3D scenes.
In practice, this is not the case.
With cubic patches of side length $10\,\mathrm{cm}$ for indoor environments and $25\,\mathrm{cm}$ for outdoor environments, typical scenes contain on the order of $5{,}000$ tokens on datasets such as ScanNet~\cite{dai2017scannet} and nuScenes~\cite{caesar2020nuscenes}.
At this scale, fused attention implementations~\cite{dao2023flashattention2,shah2024flashattention3} make full global attention practical for both training and inference.
In fact, our larger model \ours{}-B, which uses the same Transformer encoder configuration as  ViT-B, is $2\times$ faster than the prior state-of-the-art PTv3 while using $38\%$ less memory across GPU models, despite PTv3 relying only on local attention and having fewer parameters (see Fig.~\ref{fig:efficiency}).

\begin{figure*}[t]
    \centering
    \caption{
    \textbf{Inference efficiency on ScanNet.}
    Despite using full global attention, \ours{}-S is substantially faster than MinkUNet~\cite{choy20194d} while maintaining comparable peak memory usage.
    Even the larger \ours{}-B model is $2\times$ faster than PTv3-S~\cite{wu2024ptv3} while using 38\% less memory on both A100 and H100.
    }
\begin{tikzpicture}
        \begin{axis}[
            ybar,
            bar width=7pt,
            width=0.495\textwidth,
            height=5cm,
            enlarge x limits=0.15,
            legend style={at={(0.05,0.95)}, anchor=north west, legend columns=1},
            ylabel={Latency (ms) $\downarrow$},
            symbolic x coords={MinkUNet, PTv3-S, \ours{}-S, PTv3-B, \ours{}-B},
            xtick=data,
            xtick pos=lower,  x tick label style={rotate=30, anchor=north east, inner sep=2pt},
            ymajorgrids=true,
            ymin=0, ymax=120, ytick distance=25,           ]
\addplot[fill=accentblue, draw=none] coordinates {
            (MinkUNet, 38.1) (\ours{}-S, 26.5) (PTv3-S, 74.4) (\ours{}-B, 35.3) (PTv3-B, 116.7)
        };
\addplot[fill=accentred, draw=none] coordinates {
            (MinkUNet, 22.1) (\ours{}-S, 14.4) (PTv3-S, 39.1) (\ours{}-B, 20.0) (PTv3-B, 58.2) 
        };
        \legend{A100, H100}
        \end{axis}
    \end{tikzpicture}\hspace{15pt}
\begin{tikzpicture}
        \begin{axis}[
            ybar,
            bar width=7pt,
            width=0.495\textwidth,
            height=5cm,
            enlarge x limits=0.15,
            legend style={at={(0.05,0.95)}, anchor=north west,legend columns=1},
            ylabel={Peak Memory (GB) $\downarrow$},
            symbolic x coords={MinkUNet, PTv3-S, \ours{}-S, PTv3-B, \ours{}-B},
            xtick=data,
            xtick pos=lower,  x tick label style={rotate=30, anchor=north east, inner sep=2pt},
            ymajorgrids=true,
            ymin=0, ymax=4.3,
]
\addplot[fill=accentblue, draw=none] coordinates {
            (MinkUNet, 1.4) (\ours{}-S, 1.9) (PTv3-S, 3.7) (\ours{}-B, 2.3) (PTv3-B, 4.0) 
        };
\addplot[fill=accentred, draw=none] coordinates {
            (MinkUNet, 1.9) (\ours{}-S, 1.9) (PTv3-S, 3.7) (\ours{}-B, 2.3) (PTv3-B, 4.1) 
        };
\end{axis}
    \end{tikzpicture}
    \label{fig:efficiency}
\vspace{-20pt}
\end{figure*}

While \ours{} is both simple and efficient, we find that naive training on standard 3D datasets leads to pronounced overfitting.
This is not surprising: unlike domain-specific architectures with strong inductive biases, a plain Transformer with global attention has a larger hypothesis space~\cite{cordonnier2020relationship,raghu2021vision}.
This can lead to poor generalization in data-scarce regimes, but it also allows the model to identify complex patterns when enough supervision is provided.
For instance, the original Vision Transformer~\cite{dosovitskiy2021vit} underperforms in low-data regimes, but improves significantly as data and compute increase, eventually surpassing architectures with stronger inductive biases~\cite{steiner2022vitaugreg}.
We expect a similar trend to hold in 3D, but current benchmarks do not yet provide comparable scale.
For reference, widely used 3D datasets such as ScanNet~\cite{dai2017scannet} and nuScenes~\cite{caesar2020nuscenes} are orders of magnitude smaller than even modest image datasets like ImageNet~\cite{deng2009imagenet}, let alone web-scale datasets such as LAION~\cite{schuhmann2021laion400m,schuhmann2022laion5b} that power recent foundation models~\cite{cherti2023openclip,rombach2022stablediffusion}.

To address the limited scale of current 3D datasets, we refrain from incorporating stronger priors into the architecture and instead tackle the problem on the training side by adapting data-efficient Transformer recipes from 2D vision~\cite{steiner2022vitaugreg,touvron2021deit} to 3D.
In particular, we combine strong 3D data augmentations~\cite{nekrasov2021mix3d} and regularization~\cite{szegedy2016rethinking,huang2016deep} with knowledge distillation from a convolutional teacher~\cite{touvron2021deit}.
These training techniques make \ours{} competitive with domain-specific 3D backbones under standard single-dataset training.
To further increase supervision scale, we train \ours{} jointly on multiple datasets~\cite{dai2017scannet,yeshwanthliu2023scannetpp,baruch2021arkitscenes}.
With increased data, we observe consistent gains and stronger scaling behavior compared to prior backbones, achieving new best results of 80.5 mIoU on the ScanNet test set, 41.6 mIoU on the ScanNet200 test set, as well as 82.2 mIoU on the nuScenes validation set.
Finally, we demonstrate the generality of \ours{} beyond semantic segmentation by simply replacing the commonly used U-Net~\cite{choy20194d} backbone in a standard 3D instance segmentation pipeline~\cite{kolodiazhnyi2024oneformer3d} with \ours{}.
In this setting, \ours{} again attains state-of-the-art performance, reaching 82.7 mAP50 on ScanNet, 47.5 mAP50 on ScanNet200, and 54.9 mAP50 on ScanNet++ test sets, indicating that it can serve as a general-purpose 3D backbone.

In summary, our contributions are threefold:
(i) We introduce the Volume Transformer (\ours{}), a simple and efficient 3D backbone that represents 3D scenes as sequences of volumetric patch tokens, applies a vanilla Transformer encoder with full global self-attention, and injects positional information via a 3D extension of RoPE.
(ii) We propose a data-efficient training recipe for \ours{} that combines strong 3D augmentations, regularization, and distillation from a convolutional teacher, enabling effective training in today’s limited-data regime.
(iii) We demonstrate that this design scales favorably with increased supervision, achieving strong results on indoor and outdoor semantic and instance segmentation benchmarks, while being fast and memory-efficient.

\vspace{-5pt}
\section{Related Work}
Unlike images, which lie on a regular 2D grid, 3D point clouds are unordered and irregular.
Early methods, such as PointNet and PointNet++~\cite{qi2017pointnet,qi2017pointnet++}, address these challenges by treating point clouds as sets and enforcing permutation invariance through shared MLPs and symmetric pooling operations.
To better exploit local geometric structure, subsequent works introduce continuous convolution operators defined directly in 3D space~\cite{wu2019pointconv,thomas2019kpconv,thomas2024kpconvx}.
By learning kernel functions over continuous spatial neighborhoods, these methods avoid discretization artifacts and preserve fine-grained geometric details.
However, they rely on neighbor search and kernel evaluation algorithms, making them computationally expensive and sensitive to variations in point density, which limits their applicability to large-scale scenes~\cite{guo2020survey3d}.

An alternative line of work discretizes 3D space into voxels, enabling the use of 3D convolutional layers operating on a grid.
However, representing point clouds as dense voxel grids is memory-intensive since the number of voxels scales cubically with spatial resolution.
To address this, sparse convolution engines~\cite{graham2018submanifold,choy20194d,tang2020spvnas,spconv2022} exploit the inherent sparsity of 3D data by using hash-based data structures and perform convolutions only on active voxels, achieving substantial gains in efficiency.
Building on these engines, fully convolutional 3D U-Net architectures~\cite{ronneberger2015u,choy20194d,graham2018submanifold,tang2020spvnas} have been widely adopted for 3D scene understanding tasks, spanning indoor and outdoor environments as well as semantic and instance-level tasks~\cite{lai2023spherical,zhu2021cylindrical,yilmaz24mask4former,schult2023mask3d,kolodiazhnyi2024oneformer3d}.
Despite their efficiency, they are ultimately built on local convolutional operators, so global context is typically aggregated only through network depth, pooling, and multi-scale feature fusion.

Motivated by the success of Vision Transformers~\cite{dosovitskiy2021vit}, recent works explore attention-based architectures for 3D understanding.
While global self-attention offers a flexible mechanism for modeling long-range interactions, most methods constrain attention to local neighborhoods to avoid the quadratic cost on large-scale point clouds.
Examples include octree- and window-based variants such as OctFormer~\cite{Wang2023OctFormer}, Stratified Transformer~\cite{lai2022stratified}, and Swin3D~\cite{yang2023swin3d}, which adapt core principles from 2D Transformers to 3D data.
Finally, the Point Transformer family~\cite{zhao2021ptv1,wu2022ptv2,wu2024ptv3} proposes attention operators with improved efficiency and scalability.
Point Transformer v3 (PTv3), the current state-of-the-art 3D backbone, introduces space-filling curves to serialize 3D points into 1D sequences and partitions scenes into smaller chunks according to this ordering.
This design largely preserves spatial locality while enabling efficient local self-attention.
Overall, however, despite adopting Transformer components, current 3D backbone architectures remain a hybrid design rather than a direct instantiation of the Transformer encoder.
They depend on U-Net-style multi-resolution hierarchies and use local operators with hand-designed neighborhoods.
While such inductive biases are effective in current data-limited 3D benchmarks, they may constrain scalability as supervision and compute increase.

In this work, we take a complementary perspective by adopting a vanilla Transformer architecture that closely follows the ViT design and by leveraging training strategies shown to be effective in 2D.
Our results show that architectural simplicity, when paired with appropriate training recipes, can offer a scalable and highly competitive alternative for 3D scene understanding.

In parallel to scene-scale 3D perception, Transformers have also been studied for object-centric point clouds~\cite{guo2021pct,pang2022pointmae,yu2022pointbert,knaebel2023point2vec}, including shape classification and part segmentation, most commonly on synthetic benchmarks such as ModelNet40~\cite{wu2015modelnet} and ShapeNet~\cite{chang2015shapenet}.
However, these single-object settings differ fundamentally from large-scale scene understanding.
Object-level benchmarks provide isolated, canonicalized shapes with a fixed and relatively small number of points (e.g., 1{,}024 or 2{,}048), and the geometry is typically clean and uniformly sampled.
In contrast, 3D scene understanding requires handling point clouds that are orders of magnitude larger, with substantial density variation, clutter, and occlusions, and semantic reasoning that spans entire rooms or outdoor environments.
As a result, object-level Transformer architectures and training regimes do not directly transfer to full-scene 3D understanding.
\section{Method}
\begin{figure}[t]
    \centering
    \includegraphics[width=\textwidth,clip]{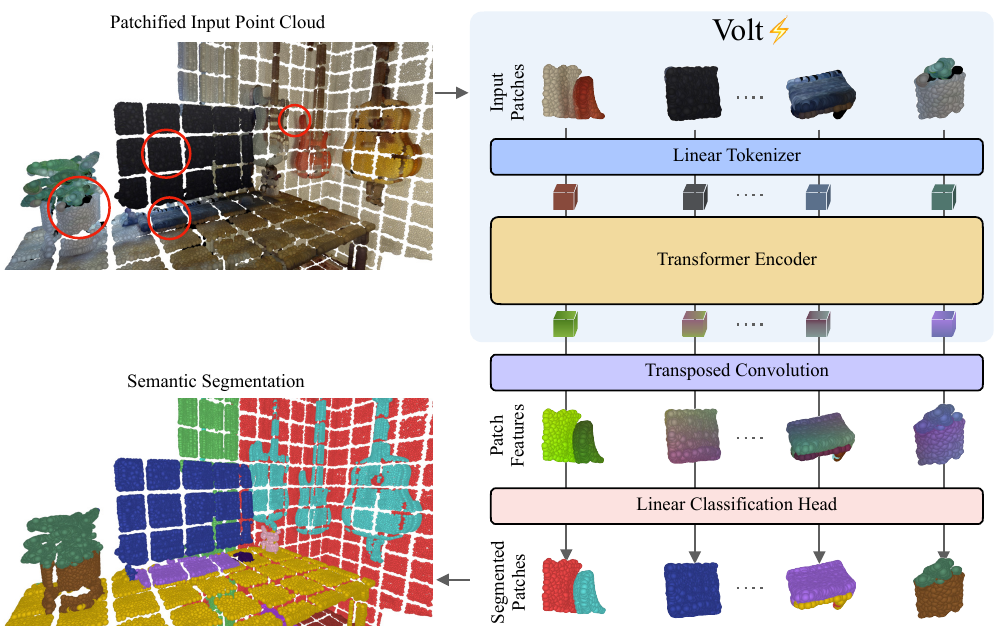}
    \caption{
    \textbf{\ours{} architecture.}
    The input 3D scene is partitioned into non-overlapping volumetric patches, and each patch is embedded into a token with a linear tokenizer. The resulting token sequence is processed by a Transformer encoder with global attention.
    The latent tokens are then upsampled back to the voxel resolution with a single transposed convolution and mapped to semantic predictions by a linear classification head.
    }
    \label{fig:method}
    \vspace{-10pt}
\end{figure}
We present the \emph{Volume Transformer} (\ours{}), a plain Transformer backbone for 3D scene understanding.
In contrast to current 3D architectures that rely on domain-specific modules~\cite{choy20194d,wu2024ptv3,Wang2023OctFormer}, \ours{} closely follows ViT~\cite{dosovitskiy2021vit} and makes a few targeted modifications for 3D data.
An overview is shown in Fig.~\ref{fig:method}.
Given a 3D scene, \ours{} partitions the scene into non-overlapping cubic patches, tokenizes them, and processes the resulting token sequence with a standard Transformer encoder using global self-attention.
The encoder outputs a set of latent tokens forming a volumetric feature map that can be used across downstream 3D tasks.
In this work, we attach a lightweight decoder to predict per-point logits for semantic segmentation and a Transformer decoder for instance segmentation.
Since \ours{} relies only on a vanilla Transformer encoder and linear layers, it can be \textit{implemented purely in PyTorch}~\cite{paszke2019pytorch} without specialized sparse-convolution engines.
Moreover, it is fully compatible with highly optimized attention kernels such as FlashAttention~\cite{dao2022flashattention,dao2023flashattention2,shah2024flashattention3} and can directly benefit from ongoing advances in the broader Transformer ecosystem.

\subsection{\ours{} Architecture}
\PARbegin{Tokenization.}
Let $\mathcal{X}=\{(\mathbf{p}_n,\mathbf{f}_n)\}_{n=1}^{N}$ denote an input point cloud with $N$ points, where each point has a 3D coordinate $\mathbf{p}_n\in\mathbb{R}^3$ and a $C$-dimensional feature vector $\mathbf{f}_n\in\mathbb{R}^{C}$ (e.g., color).
A naive application of a Transformer would treat each point as a token, yielding a sequence of length $N$ and a quadratic attention cost of $\mathcal{O}(N^2)$.
This is infeasible for typical scene-scale inputs where $N> 100{,}000$.
Instead, \ours{} first voxelizes the point cloud $\mathcal{X}$ with voxel size $\delta$ to impose a regular grid structure.
For each occupied voxel, a single representative point is selected, resulting in a sparse voxel set $\mathcal{X}_v=\{(\mathbf{z}_m,\tilde{\mathbf{f}}_m)\}_{m=1}^{M}$ where $M \leq N$, $\mathbf{z}_m\in\mathbb{Z}^3$ is the integer voxel coordinate, and $\tilde{\mathbf{f}}_m\in\mathbb{R}^{C}$ is the feature vector of the representative point in that voxel.
The sparse voxel grid is then partitioned into non-overlapping cubic patches of $P\!\times\!P\!\times\!P$ voxels.
Each patch with at least one occupied voxel, i.e., each non-empty patch, is densified by filling empty voxels with zero features while completely empty patches are discarded.
Then, each densified patch of shape $P\!\times\!P\!\times\!P$ is flattened, yielding a patch vector $\mathbf{u}_t\in\mathbb{R}^{P^3 C}$, where $t\in\{1,\dots,T\}$ are patches indices.
Finally, each patch vector $\mathbf{u}_t$ is projected to the Transformer embedding dimension $D$ using a shared learnable linear mapping, producing a token $\mathbf{x}_t \in \mathbb{R}^{D}$.
Together, these form the token sequence $\mathbf{X}=[\mathbf{x}_1,\dots,\mathbf{x}_T]$ processed by the Transformer encoder.
Since \ours{} produces a single token for each non-empty patch, the number of tokens $T$ depends on the scene size, point density, and spatial distribution of occupied voxels.
Thus, unlike in ViTs with fixed-size images, the token sequence length $T$ varies across inputs.

Overall, the tokenization step mirrors the patch embedding step in ViT: the input is split into non-overlapping regular patches, each patch is flattened and projected to the Transformer dimension.
In essence, it aggregates neighboring points and their low-dimensional features into a token, while simultaneously downsampling the scene to keep the sequence length manageable.
Compared to kNN-based grouping strategies, which are both density-dependent and can become a bottleneck at the scene scale~\cite{wu2022ptv2,wu2024ptv3}, grid-based patchification followed by a linear layer provides a simple and efficient way to form tokens.

\PAR{Transformer Encoder.}
The token sequence $\mathbf{X}$ is processed by a \emph{standard} Transformer encoder consisting of a stack of $L$ identical Transformer blocks.
Each block contains a multi-head self-attention layer and an MLP layer with residual skip connections.
Following common practice, we apply LayerNorm~\cite{ba2016layernorm} before both the attention and MLP layers.
There are no hierarchical stages, no local windows, and no domain-specific grouping strategies, but only global attention and MLPs.
To make full global self-attention practical, we use FlashAttention~\cite{dao2022flashattention,dao2023flashattention2,shah2024flashattention3}, which provides a fused and memory-efficient attention kernel for variable-length sequences.
This allows \ours{} to retain full-scene interactions while directly leveraging highly optimized, general-purpose attention implementations, in contrast to other structured attention variants that typically rely on dedicated kernels and custom indexing logic~\cite{wu2022ptv2,Wang2023OctFormer}.

\PAR{Positional Encodings.}
Injecting positional information is essential for a permutation-equivariant Transformer, and it is particularly critical in 3D, where semantics are largely determined by geometry and spatial occupancy.
Conventional learned absolute positional embeddings are ill-suited to sparse, spatially unbounded 3D scenes.
In contrast, RoPE~\cite{su2024roformer}, which has become a standard choice in modern LLMs~\cite{grattafiori2024llama3,team2025gemma3} and ViTs~\cite{simeoni2025dinov3}, is well-suited for 3D data.
RoPE injects position information by rotating the query and key vectors before scaled dot-product attention, so attention scores depend on \emph{relative} positional offsets.
It supports variable sequence lengths, generalizes beyond positions seen during training, and is compatible with fused attention kernels.

Inspired by 2D extensions of RoPE for ViTs~\cite{fang2024eva,heo2024ropevit}, we generalize RoPE to 3D, similar to concurrent work LitePT~\cite{yue2025litept}, by factorizing each query and key vector across the three spatial axes.
Let $\mathbf{q}, \mathbf{k} \in \mathbb{R}^{D_h}$ denote the per-head query and key vectors of a token at position $\mathbf{p} = (p_x, p_y, p_z)$.
We decompose $\mathbf{q}$ and $\mathbf{k}$ into axis-specific components, i.e., $\mathbf{q} = [\mathbf{q}_x, \mathbf{q}_y, \mathbf{q}_z]$ and $\mathbf{k} = [\mathbf{k}_x, \mathbf{k}_y, \mathbf{k}_z]$, and apply RoPE independently along each axis:
\begin{equation}
\begin{aligned}
\tilde{\mathbf{q}} = \mathrm{concat}\!\left(
\mathcal{R}_{\Theta}(p_x)\mathbf{q}_x,\;
\mathcal{R}_{\Theta}(p_y)\mathbf{q}_y,\;
\mathcal{R}_{\Theta}(p_z)\mathbf{q}_z
\right), \\
\tilde{\mathbf{k}} = \mathrm{concat}\!\left(
\mathcal{R}_{\Theta}(p_x)\mathbf{k}_x,\;
\mathcal{R}_{\Theta}(p_y)\mathbf{k}_y,\;
\mathcal{R}_{\Theta}(p_z)\mathbf{k}_z
\right),
\end{aligned}
\end{equation}
where $\mathcal{R}_{\Theta}(\cdot)$ denotes the standard block-diagonal rotary transformation~\cite{su2024roformer} parameterized by frequencies $\Theta$.
We use the discrete 3D token indices obtained after patchification as positions $\mathbf{p}$.
These indices are proportional to the underlying 3D coordinates up to the constant scale factor $\delta P$, preserving metric consistency across scenes and enabling the model to exploit globally meaningful spatial relationships.
In addition, we allocate the query and key dimensions asymmetrically across axes, assigning more capacity to $x$ and $y$ than to the gravity-aligned $z$ axis, reflecting the larger spatial extent of typical scenes in the horizontal plane.

\subsection{Lightweight Decoder}
3D scene understanding tasks require per-point predictions, whereas \ours{} operates on patch tokens.
Therefore, after processing the tokens through Transformer layers, we upsample them back to the original voxel resolution using an operation equivalent to a single transposed convolution.
We deliberately avoid deeper convolutional decoders to minimize overhead and reduce the risk of overfitting, since the Transformer backbone already provides strong representation capacity.
With kernel size equal to the patch size $P$, this layer serves as the natural inverse of tokenization.
In practice, we implement this operation as a linear layer followed by a voxel-shuffling operation to remove any dependence on sparse-convolution engines~\cite{spconv2022,choy20194d}.
After voxel shuffling, we retain features at the $M$ occupied input voxels, yielding $\mathbf{F}\in\mathbb{R}^{M\times D'}$.
For semantic segmentation, we map $\mathbf{F}$ to per-voxel class logits with a linear classifier.
For instance segmentation, we follow the common paradigm and use a Transformer decoder~\cite{sun2023spformer}.
In both cases, during evaluation, we obtain per-point predictions by assigning each point the prediction of its containing voxel.

\section{Training \ours{}}
Full global self-attention gives Transformers the capacity to model long-range interactions across an entire 3D scene, but it also makes them data-hungry.
In 3D, supervision is comparatively scarce, and we find that naively training \ours{} on standard 3D benchmarks leads to poor generalization.
To address this, we propose a rigorous training recipe inspired by data-efficient training strategies~\cite{touvron2021deit,steiner2022vitaugreg} in 2D vision, combining strong data augmentation, model regularization, and distillation to enable effective training of \ours{} from scratch.
Finally, we scale supervision via multi-dataset training to study the scaling behavior of \ours{}.
Fig.~\ref{fig:ablation} (left) summarizes the cumulative effect of our training recipe on \ours{} and PTv3~\cite{wu2024ptv3}.

\PAR{Data Augmentation.}
We apply a diverse set of augmentations tailored to 3D point clouds.
Specifically, scene mixing~\cite{nekrasov2021mix3d}, random cropping, and instance-level transformations alter the global scene context, whereas elastic distortions as well as random scaling, translation, rotation, and flipping perturb local geometry.
Together, these augmentations increase data diversity and encourage invariance to local geometric perturbations.

\PAR{Regularization.}
To mitigate overfitting, we apply established regularization techniques used for training Transformers~\cite{oquab2023dinov2,simeoni2025dinov3,touvron2021deit}.
We apply \emph{DropPath}~\cite{huang2016deep}, also known as stochastic depth, with a relatively high drop rate, randomly dropping the residual branches of the attention and MLP sublayers during training.
The drop rate is increased linearly with depth, so earlier layers are retained more frequently, which stabilizes optimization in deep encoders.
We further use AdamW with relatively strong \emph{weight decay} to discourage overly large weights.
At the loss level, we apply \emph{label smoothing}~\cite{szegedy2016rethinking} to prevent over-confident predictions.
Together, these choices form a simple but effective regularization recipe, consistent with best practices for ViT training.

\newcommand{\gTwo}[1]{\textcolor{mutedteal!60!white}{#1}}
\newcommand{\gFour}[1]{\textcolor{PineGreen}{#1}}
\newcommand{\gain}[1]{{\scriptsize\,(+#1)}}
\newcommand{\nogain}{{\scriptsize\,(+0.0)}}
\newcommand{\loss}[1]{{\scriptsize\,(-#1)}}

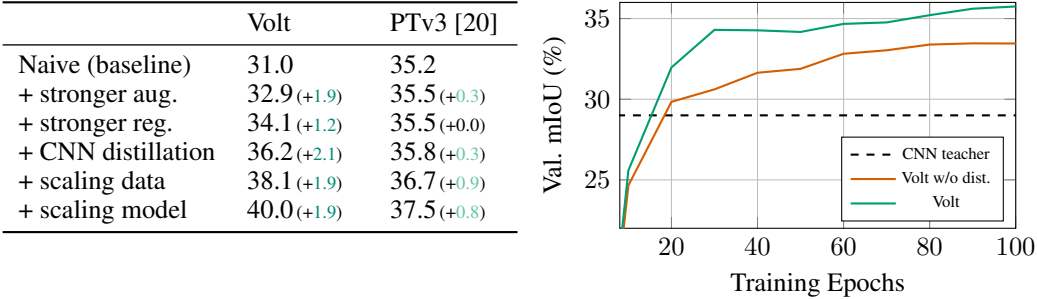
\begin{figure}[t]
  \centering
  \captionsetup{width=\textwidth}
  \caption{\textbf{Volt training recipe analysis on ScanNet200.} Left: incremental training recipe analysis. Right: training curves with and without CNN distillation.}
  \label{fig:ablation_distill}

  \begin{minipage}[t]{0.49\textwidth}
    \centering
    \label{fig:ablation}
    \begin{tabularx}{\linewidth}{l L L}
      \toprule
       & \ours{} & PTv3~\cite{wu2024ptv3} \\
      \midrule
      Naive (baseline)           & 31.0              & 35.2              \\
      + stronger aug.             & 32.9\gain{\gFour{1.9}}     & 35.5\gain{\gTwo{0.3}}     \\
      + stronger reg.          & 34.1\gain{\gFour{1.2}}     & 35.5\nogain        \\
      + CNN distillation        & 36.2\gain{\gFour{2.1}}      & 35.8\gain{\gTwo{0.3}}     \\
      + scaling data            & 38.1\gain{\gFour{1.9}}     & 36.7\gain{\gTwo{0.9}}     \\
      + scaling model           & 40.0\gain{\gFour{1.9}}     & 37.5\gain{\gTwo{0.8}}     \\
      \bottomrule
    \end{tabularx}
  \end{minipage}\hfill
  \begin{minipage}[t]{0.49\textwidth}
    \centering
    \label{fig:distill_curves}
    \captionsetup{width=\linewidth}

    \begin{tikzpicture}
      \begin{axis}[
        width=\linewidth,
        height=0.67\linewidth,
        xlabel={Training Epochs},
        ylabel={Val. mIoU (\%)},
        xmin=8, ymin=22, xmax=100, ymax=36,
        grid=both,
        legend pos=south east,
        legend style={
          font=\scriptsize,
          nodes={scale=0.9, transform shape},
        },
      ]

        \addplot[thick, dashed] table[x=epoch,y=unet] {tables/train_curves.dat};
        \addlegendentry{CNN teacher}

        \addplot[color=mutedred, thick] table[x=epoch, y expr=\thisrow{nodist}*100] {tables/train_curves.dat};
        \addlegendentry{Volt w/o dist.}

        \addplot[color=mutedteal, thick] table[x=epoch, y expr=\thisrow{dist}*100] {tables/train_curves.dat};
        \addlegendentry{Volt}

      \end{axis}
    \end{tikzpicture}
  \end{minipage}
\vspace{-15pt}
\end{figure}

\PAR{Distillation from a Convolutional Teacher.}
Prior work in 2D vision shows that distillation from CNNs transfers useful inductive biases into Transformers, improving performance especially in data-scarce regimes~\cite{abnar2020transferring,chen2022dearkd,zhao2023cumulative}.
We adopt this idea in 3D by using a 3D U-Net~\cite{choy20194d}, trained on the same data, as a teacher for \ours{}.
Note that \textit{this is not the typical distillation setting~\cite{hinton2015distilling}} where a stronger teacher guides a weaker student.
In fact, the teacher consistently underperforms \ours{} across different settings on the validation set, while still benefiting \ours{} through the distillation objective (see Fig.~\ref{fig:distill_curves}, right).

Following DeiT~\cite{touvron2021deit}, we use the teacher’s hard segmentation predictions $y_{\text{teacher}}$ as distillation targets.
Accordingly, \ours{} uses two linear classification heads on the voxel features $\mathbf{F}$: a segmentation head with weights $\mathbf{W}_{\text{seg}}$ supervised by the ground truth $y_{\text{gt}}$, and a distillation head with weights $\mathbf{W}_{\text{distill}}$ supervised by the teacher predictions $y_{\text{teacher}}$.
The loss function is defined as:
\begin{equation}
\mathcal{L}
= 0.5\,\mathcal{L}_{\text{seg}}\!\left(\mathbf{W}_{\text{seg}}^T \mathbf{F},\, y_{\text{gt}}\right)
+ 0.5\,\mathcal{L}_{\text{seg}}\!\left(\mathbf{W}_{\text{distill}}^T \mathbf{F},\, y_{\text{teacher}}\right).
\end{equation}
This joint loss function allows the model to benefit from the teacher’s inductive biases while remaining directly optimized for the semantic segmentation objective.
Moreover, because the teacher receives the same augmented inputs as the student, its predictions vary across different augmentations of the same scene, providing an additional source of regularization.
After training, the teacher is discarded, so distillation introduces no additional cost during inference.

\PAR{Scaling Data.}
As a standard Transformer with minimal inductive biases, \ours{} is expected to benefit more from additional data than domain-specific architectures~\cite{choy20194d,wu2024ptv3}, mirroring the strong scaling behavior observed in ViTs~\cite{dosovitskiy2021vit}.
To investigate this and better assess the potential of \ours{} on currently available 3D data, we train it jointly across multiple 3D datasets.
We group data by domain, performing joint training on indoor RGB-D datasets and, separately, on outdoor LiDAR datasets following standard practice~\cite{ji2025arkit,wu2024ppt,knaebel2025ditr}.
Notably, each dataset is collected with a different sensor and reconstructed via different pipelines, leading to substantial variation in point density, coverage, and noise, making cross-dataset transfer challenging~\cite{wu2024ppt}.
Nonetheless, joint training increases both the amount of supervision and the diversity of the training data, offering a glimpse of how \ours{} scales with data size.
To account for the different semantic label spaces across datasets, we use a per-dataset linear classification head while sharing the rest of the backbone.

\PAR{Training Recipe Results.}
When evaluating the impact of this training recipe in Fig.~\ref{fig:ablation}, we find that PTv3 is stronger under naive training (35.2 vs.\ 31.0 mIoU), but \ours{} benefits substantially more from the data-efficient training recipe and increased supervision.
Across the same sequence of upgrades, \ours{} improves by +9.0 mIoU (31.0$\rightarrow$40.0), whereas PTv3 gains only +2.3 mIoU (35.2$\rightarrow$37.5).
Notably, stronger regularization and CNN distillation improve \ours{} significantly more than PTv3, suggesting that they mitigate the effects of weaker inductive biases and the resulting overfitting under limited supervision.
This trend is further supported by Fig.~\ref{fig:scaling}, which evaluates segmentation accuracy as the amount of training data increases. 
Under naive training, \ours{} performs poorly in the low-data regime but improves rapidly as the training set grows, revealing substantial headroom with scale.
Applying the data-efficient training recipe significantly improves performance, making \ours{}-S consistently better than the $5\times$ larger PTv3-B model across data scales. 
At the largest supervision scale, improvements for \ours{}-S begin to saturate, suggesting performance becomes capacity-limited with $23.7$M parameters.
Scaling the backbone to \ours{}-B alleviates this saturation: \ours{}-B continues to benefit from additional supervision and opens a clear gap under multi-dataset training.
Overall, these results suggest that, as supervision increases, hand-crafted architectural priors become less critical and may even introduce unnecessary overhead, while \ours{} can instead learn domain-specific 3D patterns directly from data.

\colorlet{voltLight}{mutedteal!60!white}
\colorlet{voltMid}{mutedteal}
\colorlet{voltDark}{mutedteal!75!black}

\colorlet{ptv3Light}{mutedred!50!white}
\colorlet{ptv3Mid}{mutedred}
\colorlet{ptv3Dark}{mutedred!70!black}

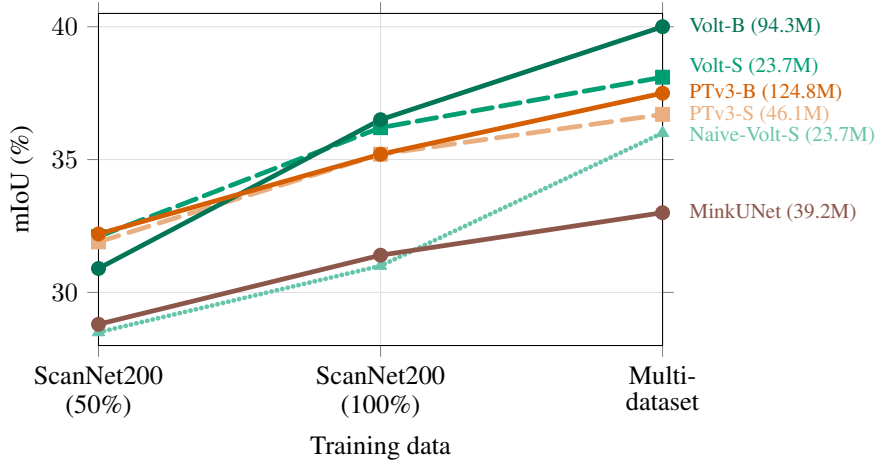
\begin{figure}[t]
\centering
\hspace*{0.1\textwidth}
\begin{tikzpicture}
\pgfplotsset{
  baseplot/.style={
    line width=1.8pt,
    line cap=round,
    line join=round,
    mark size=2.6pt,
    mark options={solid, line width=0.45pt},
  },
minkStyle/.style={baseplot, color=mutedbrown},
voltNaiveStyle/.style={baseplot, color=voltLight, dash pattern=on 0.1pt off 2.0pt}, voltSStyle/.style    ={baseplot, color=voltMid,  dash pattern=on 7.5pt off 3.2pt },
  voltBStyle/.style    ={baseplot, color=voltDark, solid },
ptv3SStyle/.style    ={baseplot, color=ptv3Light,   dash pattern=on 8pt off 4pt},
  ptv3BStyle/.style    ={baseplot, color=ptv3Mid,  solid}
}
\begin{axis}[
  width=0.65\textwidth,
  height=6cm,
  grid=both,
  xlabel={Training data},
  ylabel={mIoU (\%)},
  xmode=log,
  log basis x=10,
  xmin=50, xmax=200,
  xtick={50,100,200},
  xticklabels={ScanNet200\\(50\%),ScanNet200\\(100\%),\shortstack{Multi-\\dataset}},
  x tick label style={rotate=0, align=center},
  ymin=28.0, ymax=40.5,
  legend style={
    at={(1.02,0.50)}, anchor=west,
    draw=none,
    fill=white,
    fill opacity=0.2,
    text opacity=1,
    font=\scriptsize,
    inner sep=2pt,
    /tikz/every even column/.append style={column sep=6pt},
  },
  legend columns=1,
  tick align=outside,
  legend cell align=left,
  legend image post style={line cap=round, xscale=1.25},
  major grid style={draw=black!12},
  minor grid style={draw=black!6},
  clip=false,
]
\addplot+[voltNaiveStyle, mark=triangle*]
  table[x=x,y=oursS] {figures/scaling_data.dat}
  node[pos=1, anchor=west, font=\footnotesize, xshift=6pt, yshift=-1pt]
  {Naive-\protect\ours{}-S (23.7M)};

\addplot+[minkStyle, mark=*]
  table[x=x,y=minkunet] {figures/scaling_data.dat}
  node[pos=1, anchor=west, font=\footnotesize, xshift=6pt, yshift=0pt]
  {MinkUNet (39.2M)};

\addplot+[voltSStyle, mark=square*]
  table[x=x,y=oursSAug] {figures/scaling_data.dat}
  node[pos=1, anchor=west, font=\footnotesize, xshift=6pt, yshift=4pt]
  {\protect\ours{}-S (23.7M)};

\addplot+[ptv3SStyle, mark=square*]
  table[x=x,y=ptv3S] {figures/scaling_data.dat}
  node[pos=1, anchor=west, font=\footnotesize, xshift=6pt, yshift=0pt]
  {PTv3-S (46.1M)};

\addplot+[voltBStyle, mark=*]
  table[x=x,y=oursB] {figures/scaling_data.dat}
  node[pos=1, anchor=west, font=\footnotesize, xshift=6pt, yshift=0pt]
  {\protect\ours{}-B (87.7M)};

\addplot+[ptv3BStyle, mark=*]
  table[x=x,y=ptv3B] {figures/scaling_data.dat}
  node[pos=1, anchor=west, font=\footnotesize, xshift=6pt, yshift=0pt]
  {PTv3-B (124.8M)};
\end{axis}
\end{tikzpicture}
    \caption{
    \textbf{Scaling behavior comparison of 3D backbones.}
    Semantic segmentation mIoU on ScanNet200 when training with 50\% and 100\% of ScanNet200, and under multi-dataset joint training, also using ARKitScenes~\cite{ji2025arkit,baruch2021arkitscenes} and ScanNet++~\cite{yeshwanthliu2023scannetpp}.
    As supervision increases, \ours{} scales more favorably than prior domain-specific backbones.
    }
    \label{fig:scaling}
\end{figure}

\section{Experiments}

\PARbegin{Implementation Details.}
We introduce two variants of \ours{}, \ours{}-S and \ours{}-B, which use the same Transformer encoder configuration as ViT-S and ViT-B, respectively, and apply QKNorm~\cite{henry2020qknorm} in all attention layers.
Following common practice, we use a voxel size $\delta$ of $2\,\mathrm{cm}$ for indoor datasets and $5\,\mathrm{cm}$ for outdoor datasets.
With a patch size $P=5$, this results in cubic patches of side length $10\,\mathrm{cm}$ for indoor datasets and $25\,\mathrm{cm}$ for outdoor datasets.
For naive training, we use the same data augmentations as PTv3~\cite{wu2024ptv3}.
For our data-efficient training recipe, we additionally apply random point dropout, per-instance rotation, shift, and scaling using the provided instance masks, as well as aggressive scene mixing~\cite{nekrasov2021mix3d} with probability $0.85$.
For CNN distillation, we adopt the commonly used MinkUNet Res16UNet34C~\cite{choy20194d} as the convolutional teacher.
Following prior work~\cite{wu2024ptv3,sun2023spformer}, we use a combination of cross-entropy and Lovász loss~\cite{berman2018lovasz} for semantic segmentation, and a combination of Dice loss and cross-entropy loss for instance segmentation.
We optimize all models with AdamW~\cite{loshchilov2019adamw} using a 1-cycle~\cite{smith2019onecyclelr} learning-rate schedule, a weight decay of $0.05$, and label smoothing of $0.1$.
We maintain an exponential moving average of the model weights with decay $0.999$ for more stable evaluation.
All models are trained on 8 A100 GPUs with a global batch size of 16.
Unless stated otherwise, we use FP16 mixed precision and FlashAttention-2 for improved memory efficiency and training speed.
Following the timm library~\cite{rw2019timm}, we initialize all linear layers with a truncated normal distribution ($\sigma{=}0.02$) and set all biases to zero.

\PAR{Datasets.}
We evaluate \ours{} on a diverse suite of indoor and outdoor 3D benchmarks, following the official splits and evaluation protocols for all datasets.

For indoor scene understanding, we train and evaluate separately on the ScanNet, ScanNet200, and ScanNet++ datasets.
ScanNet~\cite{dai2017scannet} contains 1{,}613 scenes with hundreds of raw semantic annotation categories, but its standard benchmark is limited to a subset of 20 classes.
ScanNet200~\cite{rozenberszki2022scannet200} proposes to use a finer-grained label space of 200 semantic classes for the same underlying ScanNet scans, better capturing the long-tail diversity of real indoor environments.
ScanNet++~\cite{yeshwanthliu2023scannetpp} provides higher-quality 3D reconstructions and comprises 956 scenes, annotated with 100 semantic classes.
For joint training, we combine these three datasets and additionally include ARKit LabelMaker~\cite{ji2025arkit}, which augments ARKitScenes~\cite{baruch2021arkitscenes} with automatically generated semantic labels, providing 5{,}019 real scans annotated with 185 classes.
While these labels are imperfect, their scale substantially increases supervision and enables us to study the data-scaling behavior of \ours{} beyond what is possible with manually annotated data.

For outdoor evaluation, we use the nuScenes~\cite{caesar2020nuscenes}, Waymo Open Dataset~\cite{sun2020waymo}, and SemanticKITTI~\cite{behley2019semantickitti} datasets, which differ in sensor configurations and scene dynamics.
nuScenes contains 1{,}000 sequences of duration 20\,s captured with a 32-beam LiDAR and provides 16 semantic classes.
The Waymo Open Dataset is of comparable scale but uses a denser 64-beam LiDAR and contains 22 classes.
SemanticKITTI is annotated with 19 semantic classes and uses a LiDAR sensor similar to Waymo. However, it is less diverse, containing only 10 training sequences and 1 validation sequence collected in the Karlsruhe suburbs, which results in limited geographic diversity.

\subsection{Comparison with State-of-the-Art Methods}

\begin{table}[t]
	\centering
	\caption{
		\textbf{Indoor semantic segmentation results (mIoU).}
	}
	\begin{tabularx}{\linewidth}{l c Y c YY c YY c c}
		\toprule
		\multirow{2}[2]{*}{Method} && \multirow{2}[2]{*}{\#Params} &&
		\multicolumn{2}{c}{ScanNet} &&
		\multicolumn{2}{c}{ScanNet200} &&
		ScanNet++\\
		\cmidrule{5-6} \cmidrule{8-9} \cmidrule{11-11}
		&&&& Val & Test && Val & Test && Test\\
		\midrule
		ST~\cite{lai2022stratified}        && 18.8M && 74.3 & 73.7 && --   & --   && --\\
		PTv1~\cite{zhao2021ptv1}           && 11.4M && 70.6 & --   && 27.8 & --   && --\\
		PointNeXt~\cite{qian2022pointnext} && 41.6M && 71.5 & 71.2 && --   & --   && --\\
		MinkUNet~\cite{choy20194d}         && 39.2M && 72.2 & 73.6 && 25.0 & 25.3 && 45.6\\
		OctFormer~\cite{Wang2023OctFormer} && 44.0M && 75.7 & 76.6 && 32.6 & 32.6 && 46.0\\
		Swin3D~\cite{yang2023swin3d}       && 23.6M && 76.4 & --   && --   & --   && --\\
		PTv2~\cite{wu2022ptv2}             && 12.8M && 75.4 & 75.2 && 30.2 & --   && 44.5\\
		OA-CNN~\cite{peng2024oa}           && 51.5M && 76.1 & 75.6 && 32.3 & 33.3 && 47.0\\
		PTv3~\cite{wu2024ptv3}             && 46.1M && 77.5 & 77.9 && 35.2 & 37.8 && 48.8\\
		\textbf{\ours{}-S}                  && 23.7M && 77.2 & --   && 36.2 & --   && 49.3 \\
        \midrule
        \multicolumn{11}{c}{$\downarrow$~Jointly trained on multiple datasets}\\
		\midrule
        PTv3/PPT~\cite{wu2024ppt,ji2025arkit}        && 124.8M && 79.1 & 79.8 && 37.5 & 41.4 && --\\
		\textbf{\ours{}-S}                   && 23.7M && 80.2 & -- && 38.1 & -- && -- \\
        \textbf{\ours{}-B}                 && 87.7M && \textbf{80.5} & \textbf{80.5} && \textbf{40.0} & \textbf{41.6} && \textbf{49.5} \\
		\bottomrule
	\end{tabularx}
	\label{tab:indoor_semseg}
\end{table}

\begin{table}[t]
	\centering
	\caption{\textbf{Outdoor semantic segmentation results (mIoU).}
    }
	\begin{tabularx}{\linewidth}{l c Y c YY c YY c Y}
		\toprule
		\multirow{2}[2]{*}{Method} && \multirow{2}[2]{*}{\#Params} &&
        \multicolumn{2}{c}{nuScenes} && \multicolumn{2}{c}{Sem.KITTI} && Waymo\\
		\cmidrule{5-6} \cmidrule{8-9} \cmidrule{11-11}
		&&&& Val & Test && Val & Test && Val\\
		\midrule
		MinkUNet~\cite{choy20194d}            && 39.2M && 73.3& -- && 63.8& -- && 65.9\\
		SPVNAS~\cite{tang2020spvnas}          && 10.8M && 77.4& -- && 64.7& 66.4 && --\\
		Cylinder3D~\cite{zhu2021cylindrical}  && 26.1M && 76.1& 77.2&& 64.3& 67.8 && --\\
		SphereFormer~\cite{lai2023spherical}  && 32.3M && 78.4& 81.9&& 67.8& 74.8 && 69.9\\
        LSK3DNet~\cite{feng2024lsk3dnet}      && 28.8M && 80.1& -- && 70.2 & \textbf{75.6} && -- \\
		PTv2~\cite{wu2022ptv2}                && 12.8M && 80.2& 82.6&& 70.3& 72.6 && 70.6\\
		PTv3~\cite{wu2024ptv3}                && 46.1M && 80.4& 82.7&& 69.1 & 74.2 && 71.3 \\
		\textbf{\ours{}-S}                    && 23.7M && 81.0& -- && 70.5 & --  && 71.5 \\
        \midrule
        \multicolumn{11}{c}{$\downarrow$~Jointly trained on multiple datasets}\\
		\midrule
        PTv3/PPT~\cite{wu2024ppt}             && 46.3M && 81.2 & \textbf{83.0} && 72.3 & 75.5 && 72.1 \\
		\textbf{\ours{}-S}                      && 23.7M && 81.8 & -- && 72.2 & --  && 72.4\\
		\textbf{\ours{}-B}                      && 87.7M && \textbf{82.2} & 82.2 && \textbf{72.5} & 75.2  && \textbf{73.4} \\
		\bottomrule
	\end{tabularx}
	\label{tab:outdoor_semseg}
    \vspace{-15pt}
\end{table}

\PAR{Indoor Semantic Segmentation.}
Table~\ref{tab:indoor_semseg} compares \ours{} to prior work on indoor semantic segmentation.
Under single-dataset training, \ours{}-S is highly competitive and even surpasses previous approaches while being $2\times$ smaller and faster.
Next, we scale up supervision via joint training on ScanNet, ScanNet200, ScanNet++, and ARKitScenes, and compare to PPT~\cite{wu2024ppt,ji2025arkit}, which builds on the PTv3-B backbone and augments it with dataset-specific BatchNorm/LayerNorm for joint training.
We find that increasing supervision yields strong performance improvements for \ours{}-S, reaching 80.2 mIoU on ScanNet and 38.1 mIoU on ScanNet200, surpassing even the $5\times$ larger PPT trained additionally on the Structured3D dataset~\cite{zheng2020structured3d} (5 datasets in total).
We then scale up to \ours{}-B, achieving new best results across all datasets in all settings, with test-set scores of 80.5 mIoU on ScanNet, 41.6 mIoU on ScanNet200, and 49.5 mIoU on ScanNet++.
Overall, these results show that \ours{} benefits substantially from increased and more diverse supervision, and they highlight the effectiveness of a plain Transformer backbone with full scene-level self-attention for indoor 3D understanding.

Beyond segmentation accuracy, Fig.~\ref{fig:efficiency} shows that \ours{} is also efficient: using FlashAttention-2, it achieves lower inference latency on both A100 and H100 GPUs compared to MinkUNet and PTv3, despite using full global self-attention.
Even \ours{}-B remains faster than the fully convolutional MinkUNet with only a modest increase in peak GPU memory, while delivering substantially stronger segmentation accuracy.
Together, these results demonstrate that \ours{} is not only effective, but also an efficient and scalable backbone for indoor 3D understanding.

\PAR{Outdoor Semantic Segmentation.}
As summarized in Table~\ref{tab:outdoor_semseg}, \ours{} transfers cleanly to large-scale outdoor scenes, despite the sparser and irregular sampling of LiDAR compared to indoor RGB-D reconstructions.
Under single-dataset training, \ours{}-S already achieves the best performance on SemanticKITTI, nuScenes, and Waymo validation sets.
Joint training across outdoor datasets yields consistent gains, and \ours{}-S reaches state-of-the-art performance, outperforming PPT on two benchmarks and nearly matching it on the third.
Again, scaling to \ours{}-B further improves results.
Notably, these gains are achieved without introducing LiDAR-specific design choices~\cite{zhu2021cylindrical,lai2023spherical,feng2024lsk3dnet}, underscoring that the same tokenization and backbone design remains effective under substantial shifts in point density, range, and motion patterns.
Overall, \ours{} is not tied to indoor-specific priors and generalizes well to LiDAR-based perception at scale.

\begin{table}[!t]
	\centering
	\caption{
		\textbf{Indoor instance segmentation results (mAP50).}
	}
	\setlength{\tabcolsep}{3pt}
	\begin{tabularx}{\linewidth}{lccYYcYYcY}
		\toprule
		\multirow{2}[2]{*}{Method} & \multirow{2}[2]{*}{Backbone} && 
		\multicolumn{2}{c}{ScanNet}&& 
		\multicolumn{2}{c}{ScanNet200}&&
		ScanNet++\\
		\cmidrule{4-5} \cmidrule{7-8} \cmidrule{10-10}
        & & & Val& Test& & Val& Test& & Test\\
		\midrule
        SPFormer~\cite{sun2023spformer} & MinkUNet && 73.9 & 77.0 && 33.8 & -- && 43.5 \\
        Mask3D~\cite{schult2023mask3d} & MinkUNet && 73.7 & 78.0 && 37.0 & 38.8 && -- \\
        OneFormer3D~\cite{kolodiazhnyi2024oneformer3d} & MinkUNet && 78.1 & 80.1 && 40.8 & -- && 43.3 \\
        MAFT~\cite{lai2023maft} & MinkUNet && 76.5 & 78.6 && 38.2 & -- && -- \\
        QueryFormer~\cite{lu2023queryformer} & MinkUNet && 74.2 & 78.7 && 37.1 & -- && -- \\
        SphericalMask~\cite{shin2024sphericalmask} & MinkUNet && 79.9 & 81.2 && -- & -- && -- \\
        Relation3D~\cite{lu2025relation3d} & MinkUNet && 80.2 & 81.6 && 41.2 & -- && -- \\
        SGIFormer~\cite{yao2024sgiformer} & MinkUNet && \textbf{81.2} & 79.9 && 39.4 & -- && 45.7 \\
		\midrule
		SPFormer~\cite{sun2023spformer} & \textbf{\ours{}-S} && 78.1 & -- && \textbf{48.4} & -- && -- \\
		SPFormer~\cite{sun2023spformer} & \textbf{\ours{}-B} && 78.3 & \textbf{82.7} && \textbf{48.4} & \textbf{47.5} && \textbf{54.9} \\
		\bottomrule
	\end{tabularx}
	\label{tab:indoor_insseg}
\end{table}

\PAR{Indoor Instance Segmentation.}
We further evaluate \ours{} on indoor 3D instance segmentation to test whether the backbone improvements extend beyond semantic segmentation.
For a controlled comparison, we follow the established SPFormer pipeline~\cite{sun2023spformer} and keep the decoder and the losses unchanged, replacing only the original MinkUNet~\cite{choy20194d} backbone with \ours{}.
Following prior work~\cite{yao2024sgiformer,kolodiazhnyi2024oneformer3d,lai2023maft,lu2025relation3d,shin2024sphericalmask}, we initialize \ours{} from our semantic segmentation checkpoint and fine-tune for instance segmentation.
Results are reported in Table~\ref{tab:indoor_insseg}.
We find that swapping MinkUNet for \ours{}-S yields substantial gains over the SPFormer baseline, with an especially large improvement on ScanNet200 (+14.6 mAP${50}$), setting a new state of the art on the validation set.
Scaling the backbone further to \ours{}-B achieves new best results on all three test sets, reaching 82.7 mAP${50}$ on ScanNet, 47.5 mAP$_{50}$ on ScanNet200, and 54.9 mAP${50}$ on ScanNet++.
Overall, these results suggest that strengthening just the underlying 3D representation can deliver gains larger than those obtained from increasingly specialized decoder designs~\cite{kolodiazhnyi2024oneformer3d,yao2024sgiformer,lu2025relation3d}.

\PAR{Functionality Segmentation.} Finally, we evaluate \ours{} on SceneFun3D~\cite{delitzas2024scenefun3d}, a benchmark for fine-grained functional understanding of real-world indoor scenes. SceneFun3D provides high-resolution 3D reconstructions captured with Faro laser scanners and annotations of functional interactive elements, such as handles, knobs, and buttons, across nine affordance categories. The released benchmark comprises 200 training, 30 validation, and 85 hidden test scenes. We consider the functionality segmentation task, which requires jointly segmenting functional element instances and predicting their associated affordance labels. Unlike conventional instance segmentation, the target regions are often small object parts, making fine-grained 3D understanding particularly challenging. As shown in Table~\ref{tab:scenefun3d_pe} (left), \ours{}-B achieves 34.5 AP$_{50}$ on the test set, substantially outperforming previous published methods. These results demonstrate that \ours{} extends beyond conventional semantic and instance segmentation to fine-grained functional scene understanding.

\subsection{Analysis Experiments}

\PARbegin{Positional Encoding.}
As described in Sec.~3.1, we partition each attention head across the $x$, $y$, and $z$ axes and apply RoPE~\cite{su2024roformer} independently along each axis.
In our default design, we use an asymmetric allocation of $12$, $12$, and $8$ frequency pairs for $x$, $y$, and $z$, respectively.
This choice is motivated by the structure of 3D scenes, where variation along the gravity-aligned $z$ axis is often more limited than along $x$ and $y$, especially in outdoor LiDAR data.
Table~\ref{tab:scenefun3d_pe} (right) shows that this asymmetric RoPE design performs best, reaching $36.2$ mIoU on ScanNet200 and $70.5$ mIoU on SemanticKITTI.
Using a symmetric RoPE variant reduces performance to $35.4$ and $69.5$, respectively, confirming that how the head dimensions are distributed across spatial axes matters in practice.

Our default formulation computes RoPE from the discrete token indices obtained after patchification.
These indices are proportional to the original 3D coordinates up to the constant scale factor $\delta P$, preserving a consistent metric structure across scenes.
We also ablate per-scene coordinate normalization, in which token positions are independently rescaled to $[0,1]$ before computing positional encodings.
This removes the cross-scene metric consistency of the default formulation, causing the same physical distance to correspond to different normalized coordinates depending on scene extent.
As shown in Table~\ref{tab:scenefun3d_pe} (right), this degrades performance to $35.7$ mIoU on ScanNet200 and $69.3$ mIoU on SemanticKITTI.

Finally, replacing RoPE with standard Fourier positional encoding~\cite{tancik2020fourier,schult2023mask3d} leads to a clear drop to $28.6$ on ScanNet200 and $66.0$ on SemanticKITTI.
We attribute this to the fact that RoPE injects position directly into the query-key interaction, making attention scores depend on relative spatial offsets, whereas Fourier features are added to the token embeddings beforehand and must be interpreted implicitly by the network.
RoPE's relative formulation appears to provide a stronger and more natural positional inductive bias.
Overall, these results support our final design choice: RoPE is more effective than Fourier encoding, preserving metric-space positions is preferable to per-scene normalization, and an asymmetric allocation across spatial axes yields the strongest performance.

\begin{table}[t]
    \centering
    \caption{
        \textbf{Left: SceneFun3D functionality segmentation results.}
        We report results on the test split.
        \textbf{Right: Positional encoding ablation.}
        We compare our default asymmetric RoPE design against variants with per-scene coordinate normalization, a symmetric allocation across spatial axes, or standard Fourier positional encoding.
    }
    \label{tab:scenefun3d_pe}
    \setlength{\tabcolsep}{2pt}
    \begin{minipage}[t]{0.48\textwidth}
        \vspace{0pt}
        \centering
        \begin{tabular}[t]{lccc}
            \toprule
            Method & AP & AP$_{50}$ & AP$_{25}$ \\
            \midrule            LERF~\cite{kerr2023lerf,delitzas2024scenefun3d}
                & 4.8 & 12.3 & 18.1 \\
            OpenTrack3D~\cite{zhou2026opentrack3d}
                & 6.5 & 14.0 & 27.8 \\
            Mask3D-F~\cite{schult2023mask3d,delitzas2024scenefun3d}
                & 7.9 & 18.3 & 26.6 \\
            \midrule
            \ours{}-B
                & \textbf{22.7} & \textbf{34.5} & \textbf{42.6} \\
            \bottomrule
        \end{tabular}
    \end{minipage}
    \hfill
    \begin{minipage}[t]{0.49\textwidth}
        \vspace{0pt}
        \centering
        \begin{tabular}[t]{lcc}
            \toprule
            Setting & ScanNet200 & Sem.KITTI \\
            \midrule
            Asym. RoPE (\textbf{\ours{}})
                & \textbf{36.2} & \textbf{70.5} \\
            Sym. RoPE
                & 35.4 & 69.5 \\
            Asym. RoPE + norm.
                & 35.7 & 69.3 \\
            Fourier PE
                & 28.6 & 66.0 \\
            \bottomrule
        \end{tabular}
    \end{minipage}
    \vspace{-10pt}
\end{table}

\PAR{Convolutional Decoder.}
Table~\ref{tab:decoder_ablation} (left) ablates the decoder used to produce dense semantic predictions.
Without a decoder, we predict directly from patch tokens.
This variant is the fastest, but it discards fine-grained spatial detail.
Consequently, performance drops by $0.6$ mIoU on SemanticKITTI and $1.2$ mIoU on ScanNet200.
Notably, though, even without any decoder \ours{} remains competitive.
We also evaluate a larger decoder that appends two residual convolution blocks at the highest resolution after the transposed convolution.
This variant is slightly slower and even reduces accuracy, suggesting that the Transformer backbone already provides strong representations and that a heavier decoder can exacerbate overfitting under limited supervision.
Overall, a minimal decoder captures most of the benefit with only modest overhead.

\PAR{Patch Size.}
Table~\ref{tab:patch_ablation} (right) ablates the cubic patch size $P$ used for tokenization.
Using smaller patches of size $3\!\times\!3\!\times\!3$ yields the highest accuracy on both ScanNet200 (37.7 mIoU) and SemanticKITTI (71.0 mIoU), but is substantially slower due to the longer token sequence (2.5$\times$ on ScanNet200 and 2.3$\times$ on SemanticKITTI).
Conversely, a larger patch size of $7\!\times\!7\!\times\!7$ shortens the sequence and improves latency, but noticeably degrades performance, with a larger drop on ScanNet200, reflecting the sensitivity of small objects and thin structures to coarse tokenization.
Overall, $5\!\times\!5\!\times\!5$ provides a good trade-off, retaining most of the accuracy while delivering a large speedup over smaller patches.

\begin{table}[!t]
	\centering
    \caption{\textbf{Left: Conv. decoder ablation.} Having a lightweight decoder consistently boosts mIoU with modest overhead. Conversely, a larger decoder yields lower segmentation accuracy, due to increased overfitting.
    \textbf{Right: Patch size ablation.} Larger patches shorten the token sequence and improve latency but reduce segmentation accuracy. $5\!\times\!5\!\times\!5$ patches offer the best overall trade-off.}
	\begin{minipage}[t]{0.48\textwidth}
		\centering
		\setlength{\tabcolsep}{2pt}
		\begin{tabularx}{\linewidth}{lcYYcYY}
			\toprule
			\multirow{2}[2]{*}{Setting} && \multicolumn{2}{c}{ScanNet200} && \multicolumn{2}{c}{Sem.KITTI}\\
			\cmidrule{3-4} \cmidrule{6-7}
			&& mIoU & Speed && mIoU & Speed\\
			\midrule
			no decoder  && 35.0 & \textbf{26\,ms} && 69.9 & \textbf{48\,ms} \\
			\ours{}-S      && \textbf{36.2} & 27\,ms && \textbf{70.5} & 49\,ms \\
			large decoder  && 35.8 & 29\,ms && 69.5 & 50\,ms \\
			\bottomrule
		\end{tabularx}
		\label{tab:decoder_ablation}
	\end{minipage}\hfill
	\begin{minipage}[t]{0.48\textwidth}
		\centering
		\setlength{\tabcolsep}{2pt}
		\begin{tabularx}{\linewidth}{lcYYcYY}
			\toprule
			\multirow{2}[2]{*}{Patch size} && \multicolumn{2}{c}{ScanNet200} && \multicolumn{2}{c}{Sem.KITTI}\\
			\cmidrule{3-4} \cmidrule{6-7}
			&& mIoU & Speed && mIoU & Speed\\
			\midrule
			$3\!\times\!3\!\times\!3$ && \textbf{37.7} & 67\,ms && \textbf{71.0} & 113\,ms \\
			$5\!\times\!5\!\times\!5$ && 36.2 & 27\,ms && 70.5 & 49\,ms \\
			$7\!\times\!7\!\times\!7$ && 33.5 & \textbf{26\,ms} && 69.6 & \textbf{34\,ms} \\
			\bottomrule
		\end{tabularx}
		\label{tab:patch_ablation}
	\end{minipage}
    \vspace{-10pt}
\end{table}

\subsection{Qualitative Results}

\definecolor{scannet_wall}{RGB}{174,199,232}
\definecolor{scannet_floor}{RGB}{152,223,138}
\definecolor{scannet_cabinet}{RGB}{31,119,180}
\definecolor{scannet_bed}{RGB}{255,187,120}
\definecolor{scannet_chair}{RGB}{188,189,34}
\definecolor{scannet_sofa}{RGB}{140,86,75}
\definecolor{scannet_table}{RGB}{255,152,150}
\definecolor{scannet_door}{RGB}{214,39,40}
\definecolor{scannet_window}{RGB}{197,176,213}
\definecolor{scannet_bookshelf}{RGB}{148,103,189}
\definecolor{scannet_picture}{RGB}{196,156,148}
\definecolor{scannet_counter}{RGB}{23,190,207}
\definecolor{scannet_desk}{RGB}{247,182,210}
\definecolor{scannet_curtain}{RGB}{219,219,141}
\definecolor{scannet_refrigerator}{RGB}{255,127,14}
\definecolor{scannet_shower_curtain}{RGB}{158,218,229}
\definecolor{scannet_toilet}{RGB}{44,160,44}
\definecolor{scannet_sink}{RGB}{112,128,144}
\definecolor{scannet_bathtub}{RGB}{227,119,194}
\definecolor{scannet_otherfurniture}{RGB}{82,84,163}
\definecolor{scannet_ignore}{RGB}{0,0,0}

\begin{figure*}[t]
    \centering
    \setlength{\tabcolsep}{4pt}
    \begin{tabularx}{\textwidth}{@{}YYY@{}}

    \textbf{Point Cloud} & \textbf{Prediction} & \textbf{Ground Truth} \\
    [0.5ex]

    \includegraphics[width=\linewidth, trim={0 5px 0 0}, clip]{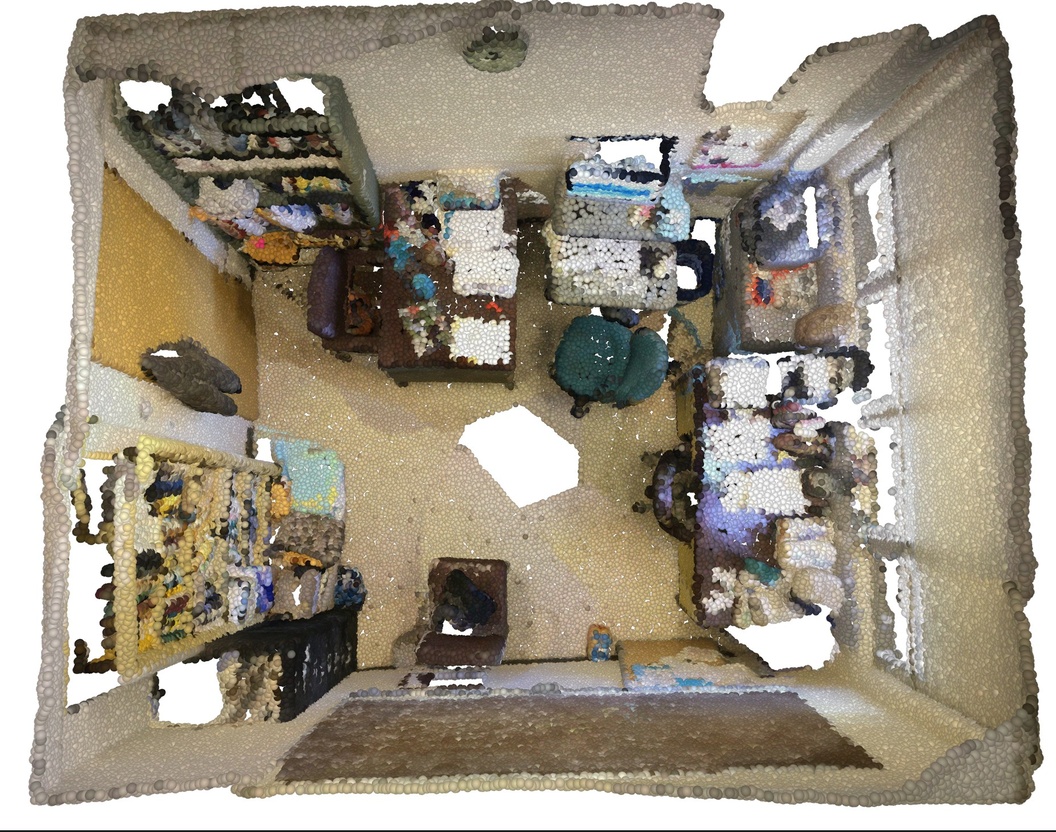}
    &
    \includegraphics[width=\linewidth, trim={0 5px 0 0}, clip]{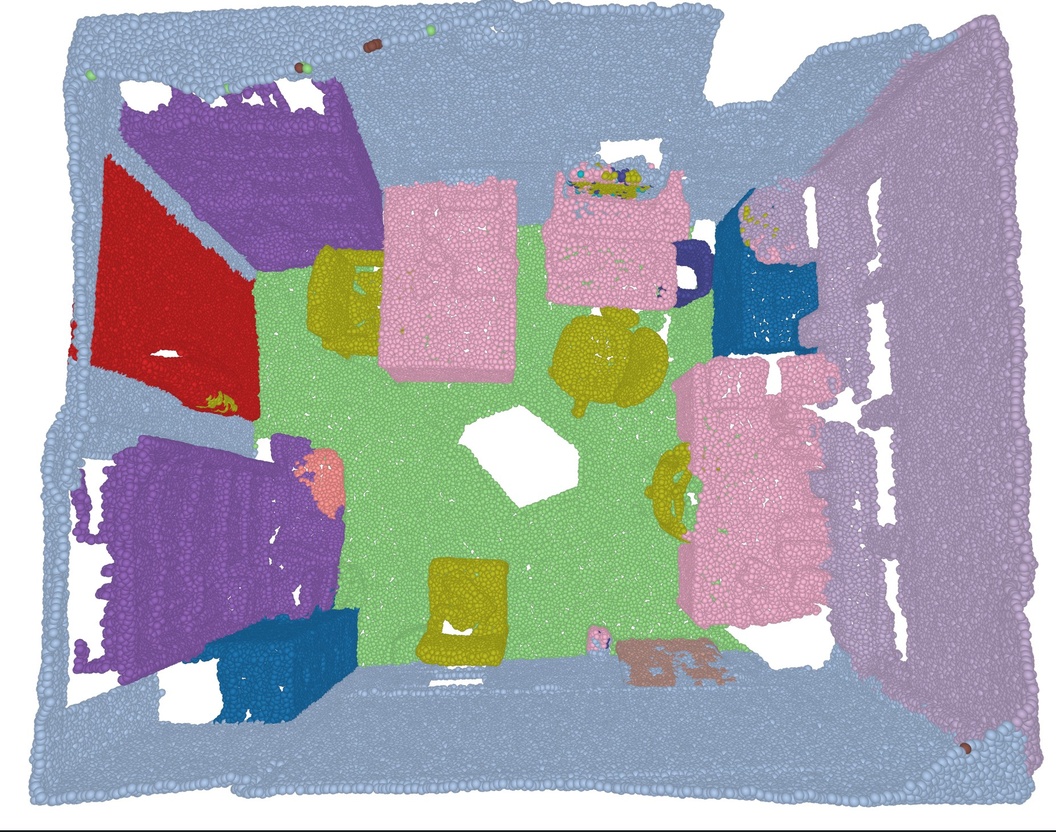}
    &
    \includegraphics[width=\linewidth, trim={0 5px 0 0}, clip]{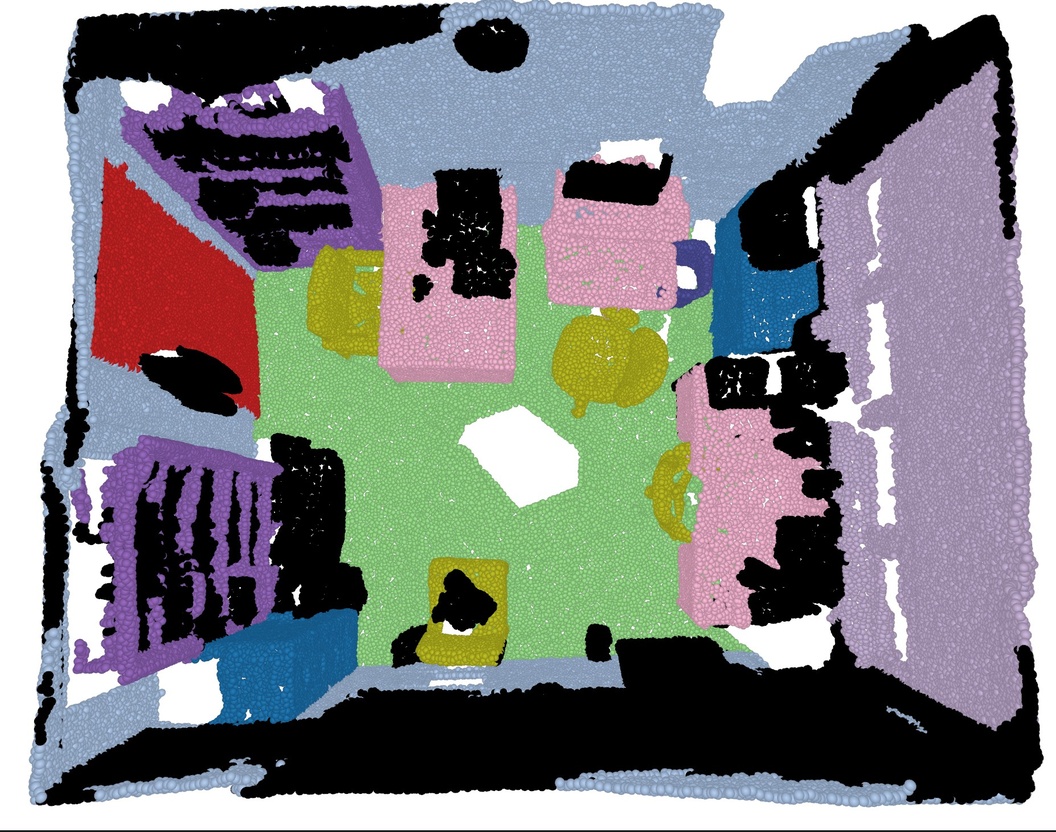}
    \\

    \includegraphics[width=\linewidth, trim={0 5px 0 0}, clip]{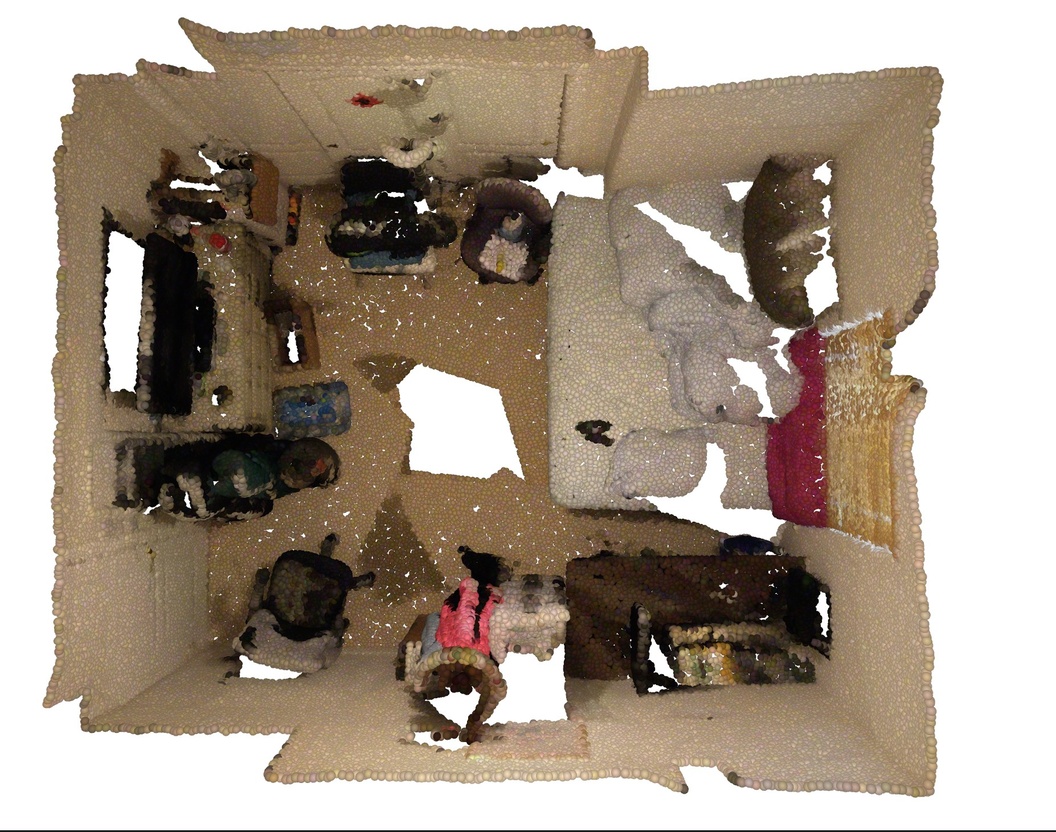}
    &
    \includegraphics[width=\linewidth, trim={0 5px 0 0}, clip]{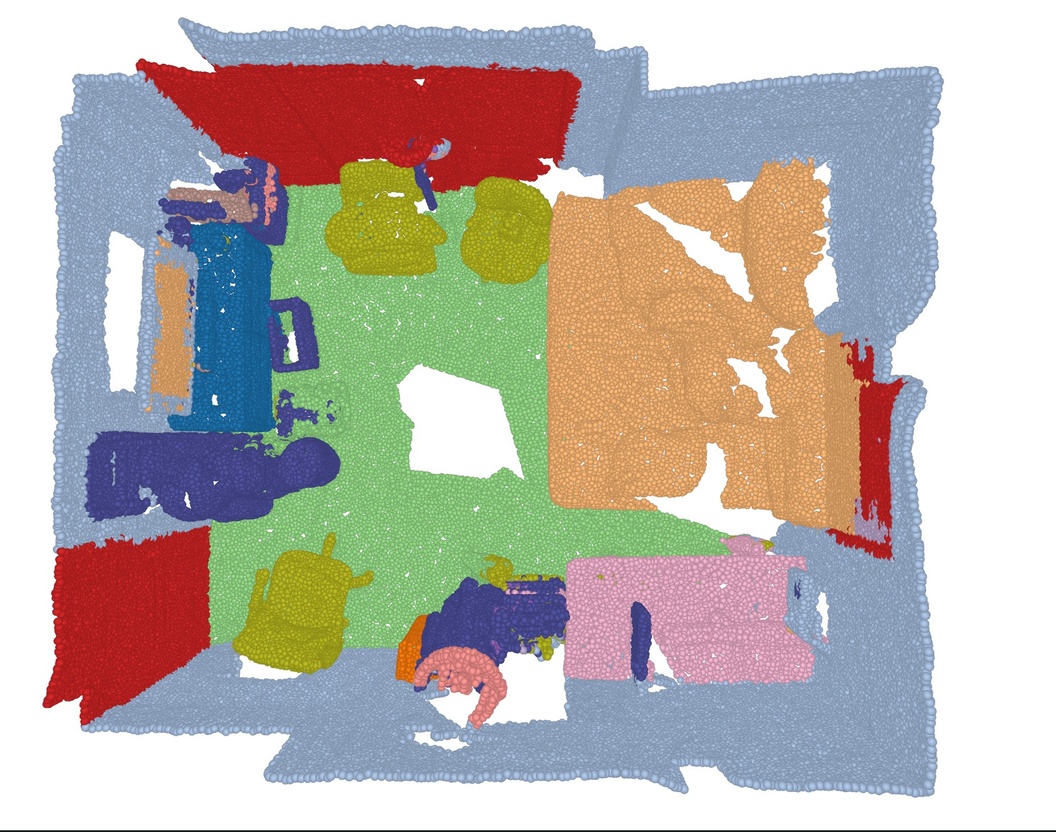}
    &
    \includegraphics[width=\linewidth, trim={0 5px 0 0}, clip]{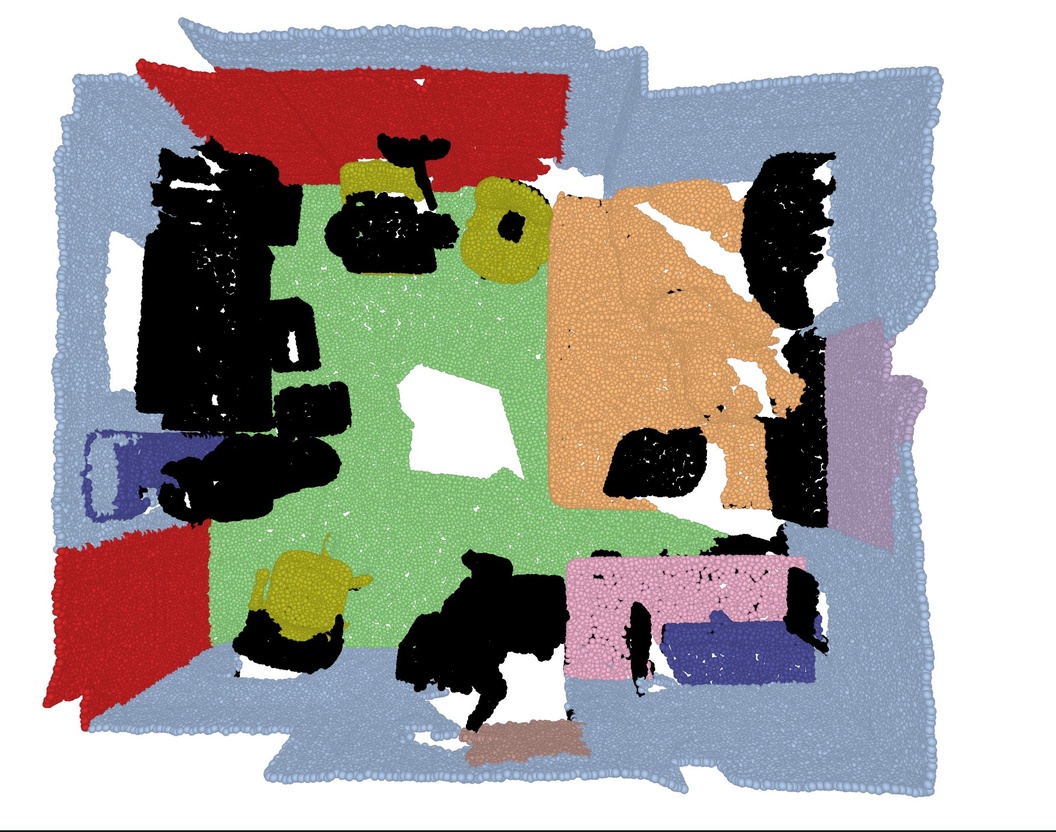}
    \\

    \includegraphics[width=\linewidth, trim={0 5px 0 0}, clip]{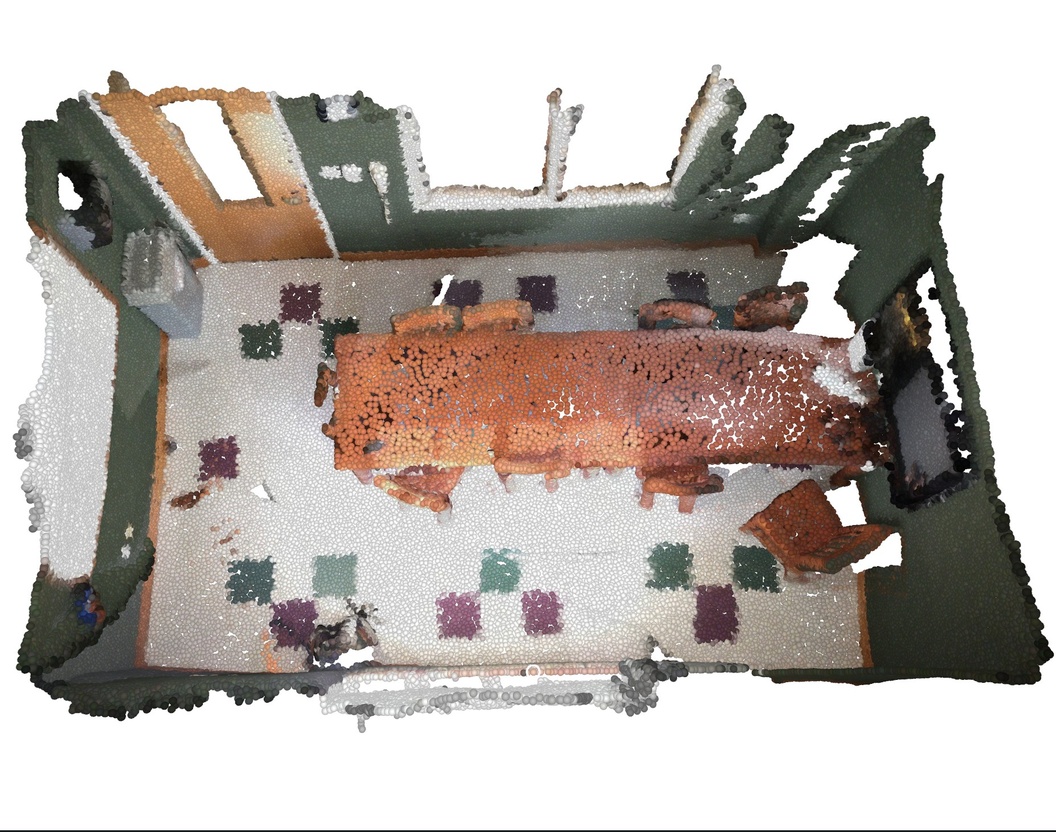}
    &
    \includegraphics[width=\linewidth, trim={0 5px 0 0}, clip]{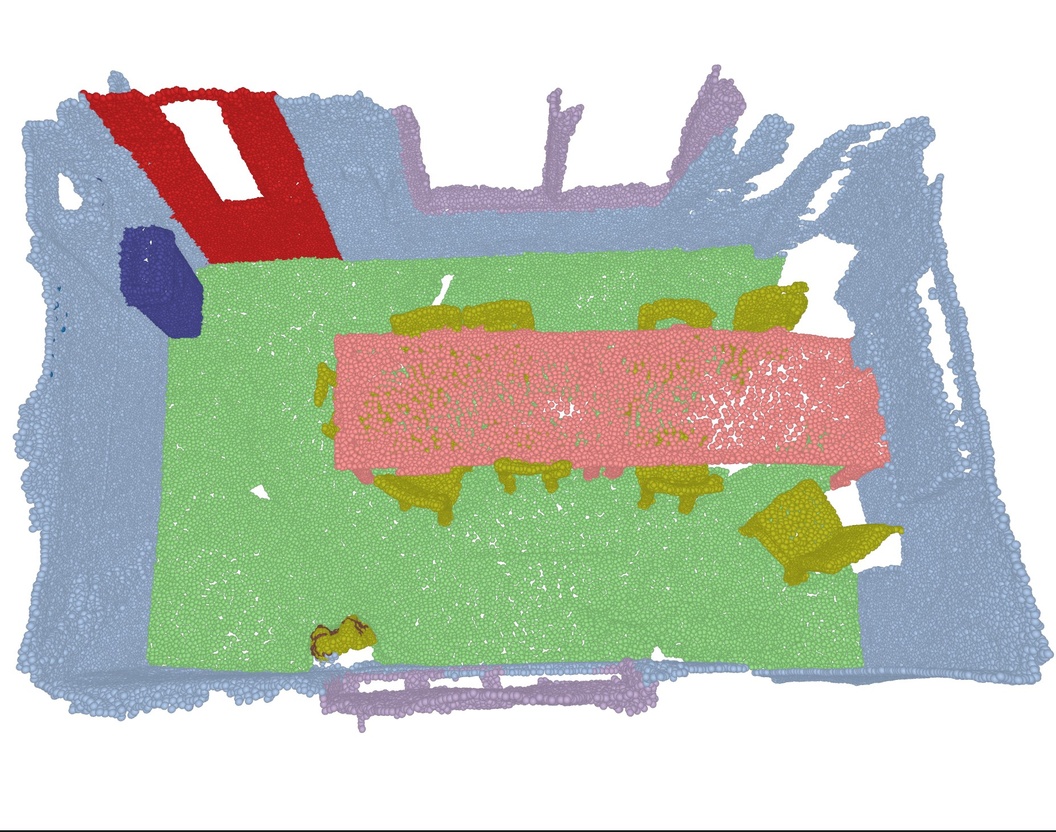}
    &
    \includegraphics[width=\linewidth, trim={0 5px 0 0}, clip]{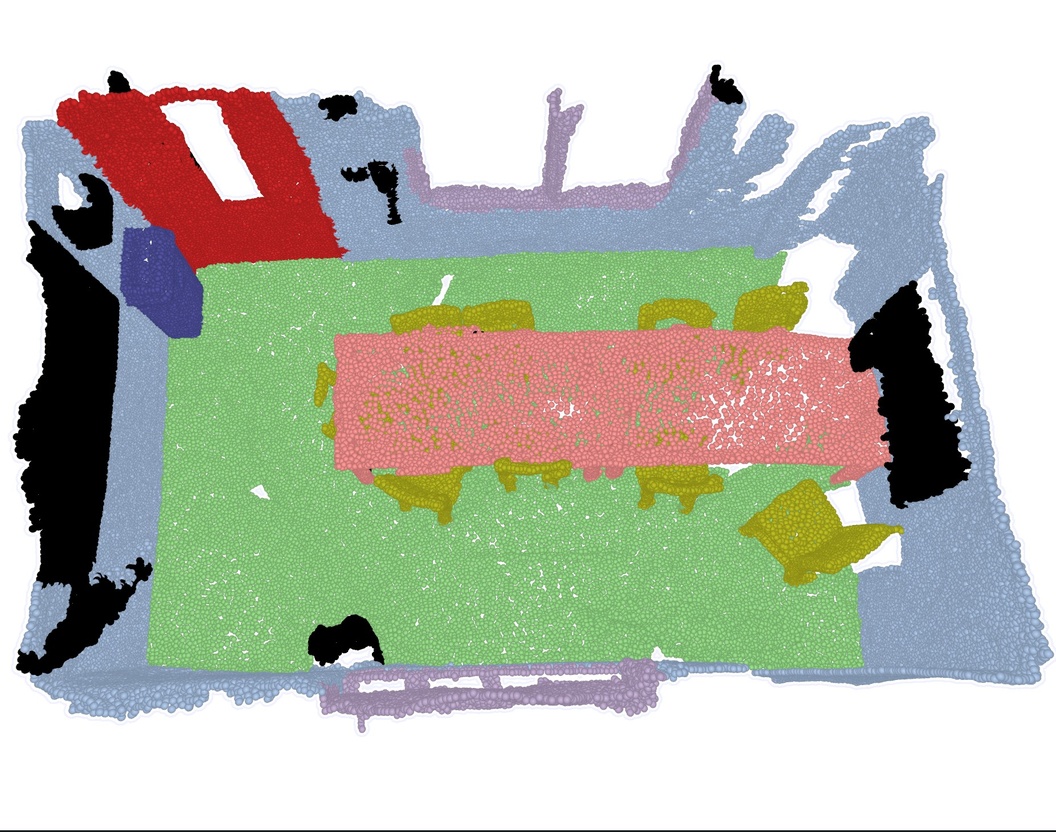}
    \\

    \includegraphics[width=\linewidth]{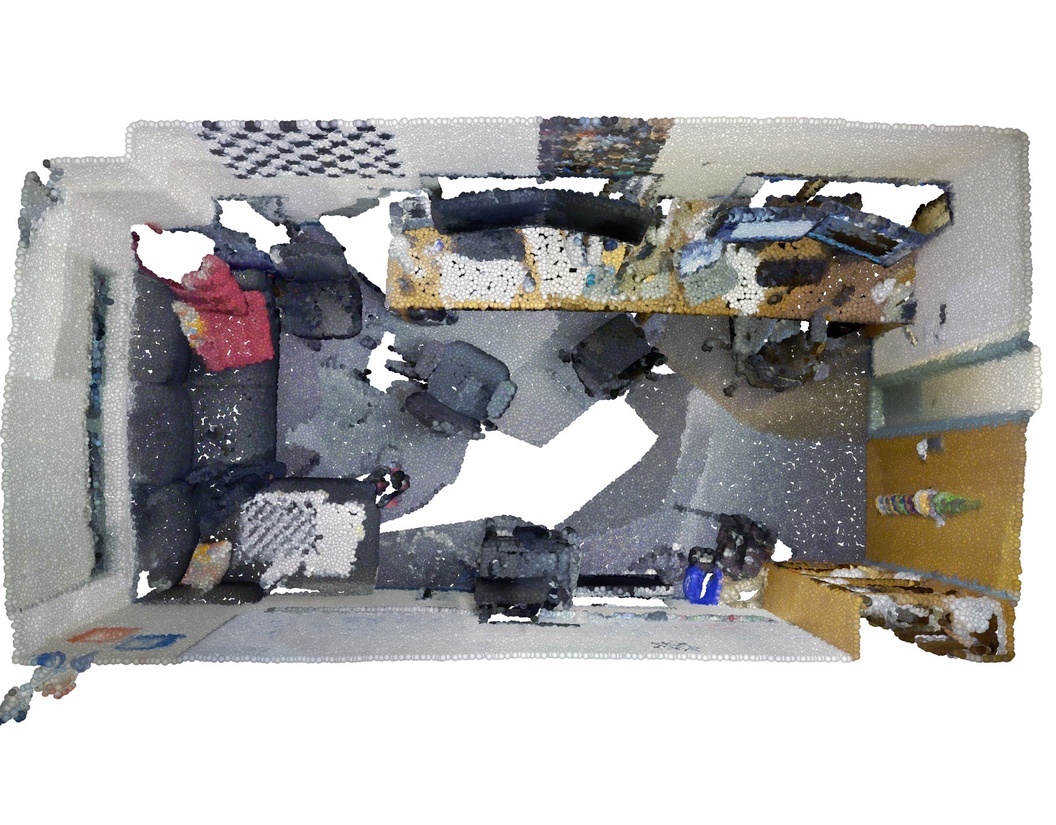}
    &
    \includegraphics[width=\linewidth]{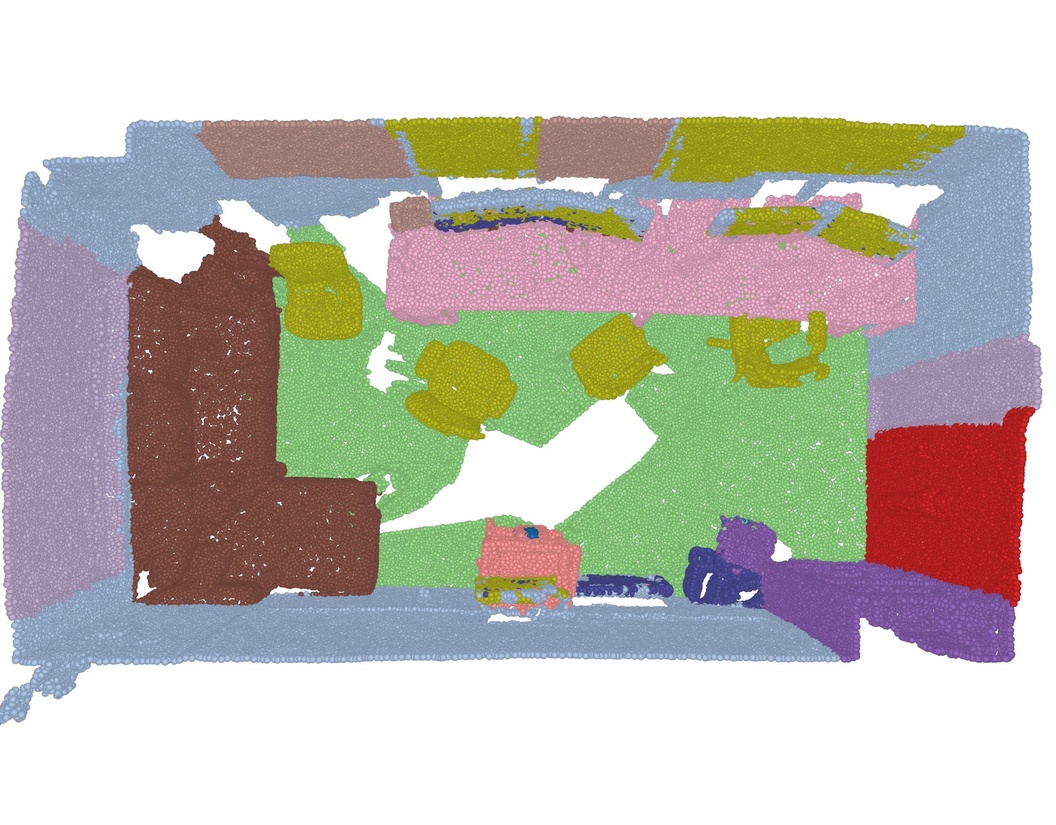}
    &
    \includegraphics[width=\linewidth]{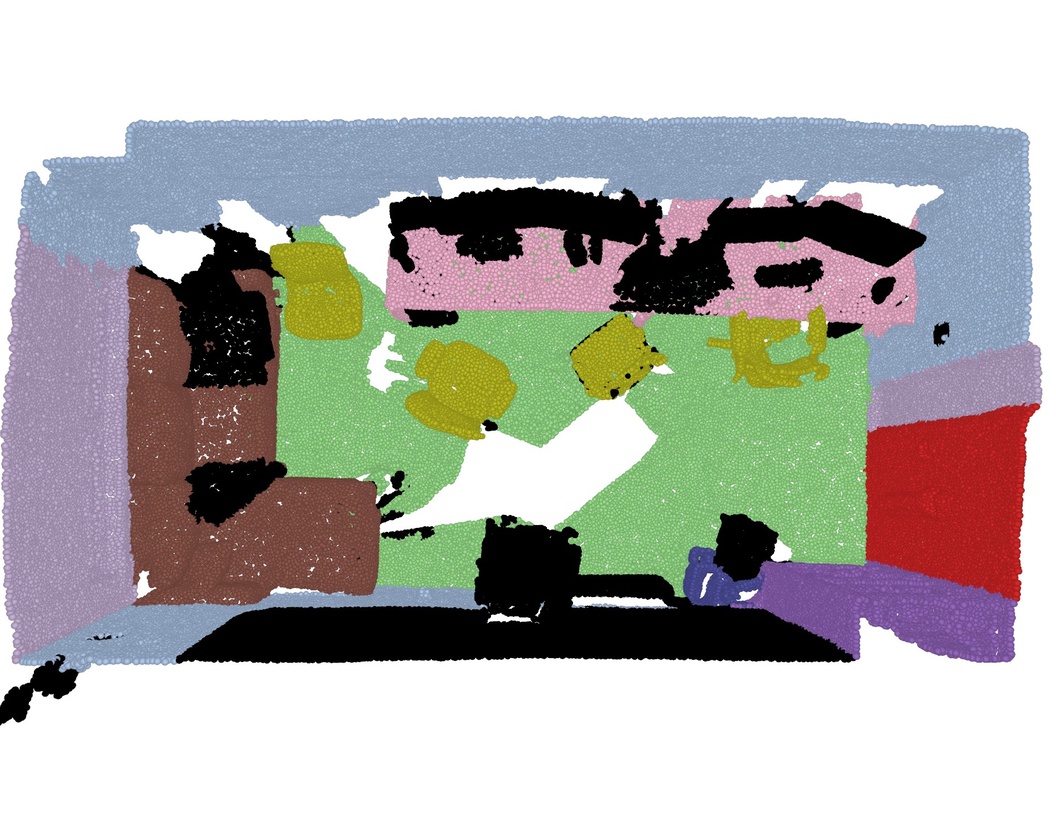}
    \\

    \includegraphics[width=\linewidth]{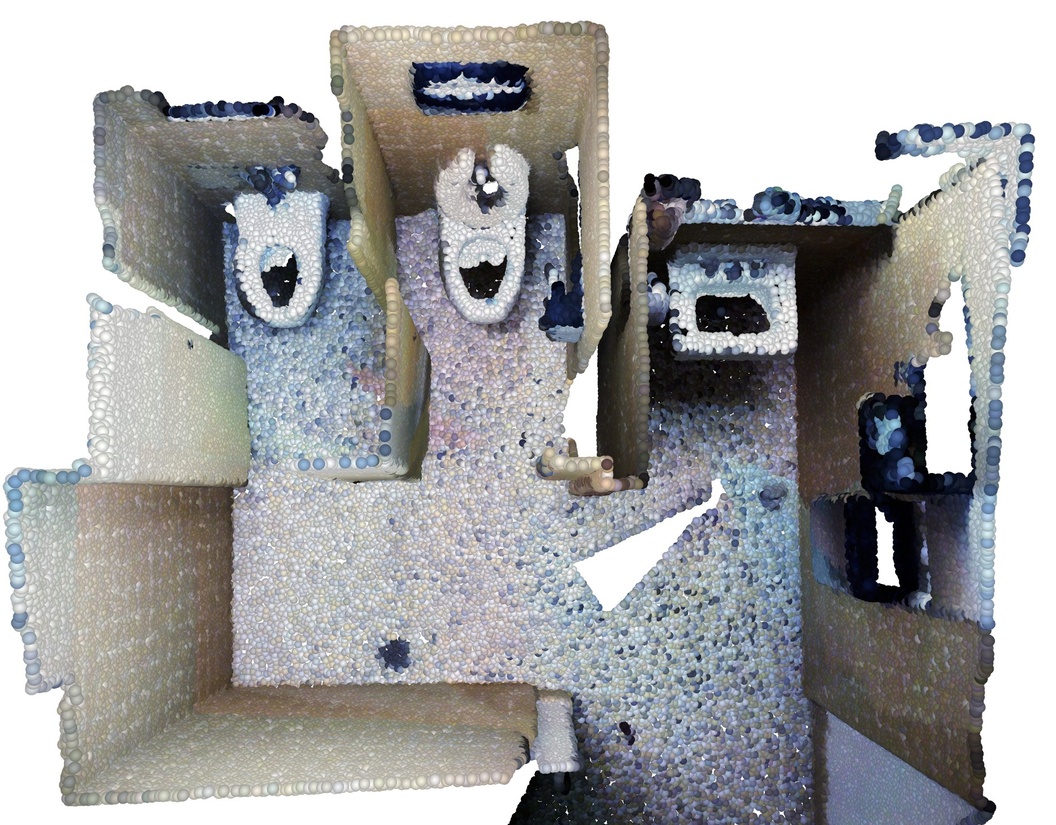}
    &
    \includegraphics[width=\linewidth]{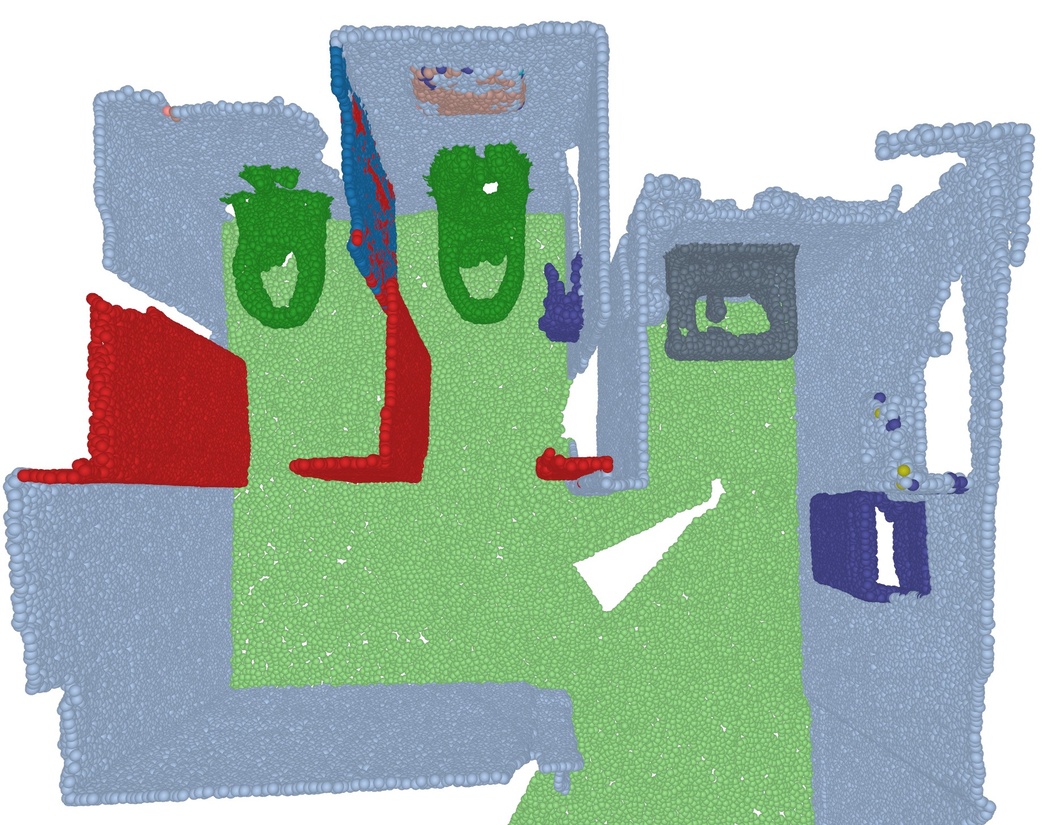}
    &
    \includegraphics[width=\linewidth]{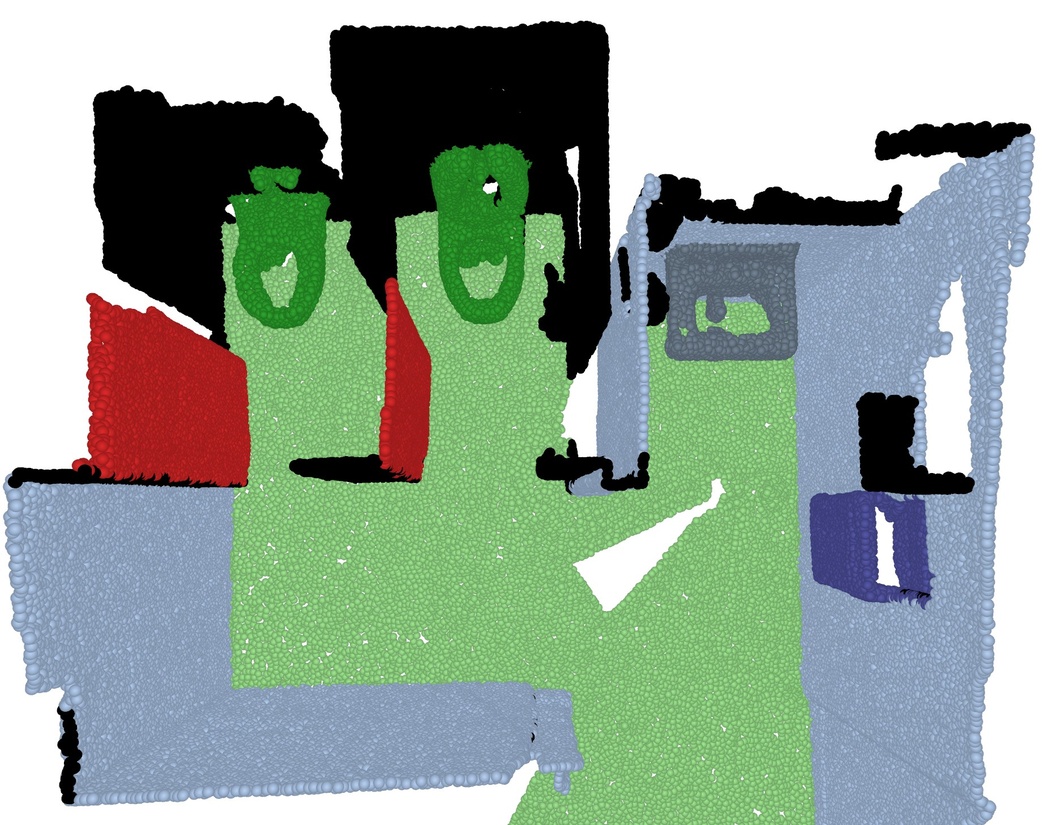}
    \\

\\[2ex]
    \multicolumn{3}{c}{\begin{minipage}{\textwidth}
    \centering
    \colorsquare{scannet_wall}{scannet_wall}\,wall \quad
    \colorsquare{scannet_floor}{scannet_floor}\,floor \quad
    \colorsquare{scannet_cabinet}{scannet_cabinet}\,cabinet \quad
    \colorsquare{scannet_bed}{scannet_bed}\,bed \quad
    \colorsquare{scannet_chair}{scannet_chair}\,chair \quad
    \colorsquare{scannet_sofa}{scannet_sofa}\,sofa \quad
    \colorsquare{scannet_table}{scannet_table}\,table \quad
    \colorsquare{scannet_door}{scannet_door}\,door \quad
    \colorsquare{scannet_window}{scannet_window}\,window \quad
    \colorsquare{scannet_bookshelf}{scannet_bookshelf}\,bookshelf \quad
    \colorsquare{scannet_picture}{scannet_picture}\,picture \quad
    \colorsquare{scannet_counter}{scannet_counter}\,counter \quad
    \colorsquare{scannet_desk}{scannet_desk}\,desk \quad
    \colorsquare{scannet_curtain}{scannet_curtain}\,curtain \quad
    \colorsquare{scannet_refrigerator}{scannet_refrigerator}\,refrigerator \quad
    \colorsquare{scannet_shower_curtain}{scannet_shower_curtain}\,shower-curtain \quad
    \colorsquare{scannet_toilet}{scannet_toilet}\,toilet \quad
    \colorsquare{scannet_sink}{scannet_sink}\,sink \quad
    \colorsquare{scannet_bathtub}{scannet_bathtub}\,bathtub \quad
    \colorsquare{scannet_otherfurniture}{scannet_otherfurniture}\,other-furniture \quad
    \colorsquare{scannet_ignore}{scannet_ignore}\,ignore
    \end{minipage}}

    \end{tabularx}
    \caption{\textbf{Qualitative semantic segmentation results on ScanNet.}}
    \label{fig:qualitative_scannet}
\end{figure*}
\definecolor{kitti_car}{RGB}{100,150,245}
\definecolor{kitti_bicycle}{RGB}{100,230,245}
\definecolor{kitti_motorcycle}{RGB}{30,60,150}
\definecolor{kitti_truck}{RGB}{80,30,180}
\definecolor{kitti_othervehicle}{RGB}{0,0,255}
\definecolor{kitti_person}{RGB}{255,30,30}
\definecolor{kitti_bicyclist}{RGB}{255,40,200}
\definecolor{kitti_motorcyclist}{RGB}{150,30,90}
\definecolor{kitti_road}{RGB}{255,0,255}
\definecolor{kitti_parking}{RGB}{255,150,255}
\definecolor{kitti_sidewalk}{RGB}{75,0,75}
\definecolor{kitti_otherground}{RGB}{175,0,75}
\definecolor{kitti_building}{RGB}{255,200,0}
\definecolor{kitti_fence}{RGB}{255,120,50}
\definecolor{kitti_vegetation}{RGB}{0,175,0}
\definecolor{kitti_trunk}{RGB}{135,60,0}
\definecolor{kitti_terrain}{RGB}{150,240,80}
\definecolor{kitti_pole}{RGB}{255,240,150}
\definecolor{kitti_trafficsign}{RGB}{255,0,0}
\definecolor{kitti_ignore}{RGB}{0,0,0}

\begin{figure*}[t]
    \centering
    \setlength{\tabcolsep}{4pt}
    \begin{tabularx}{\textwidth}{@{}YYY@{}}

    \textbf{Point Cloud} & \textbf{Prediction} & \textbf{Ground Truth} \\
    [0.5ex]

    \includegraphics[width=\linewidth]{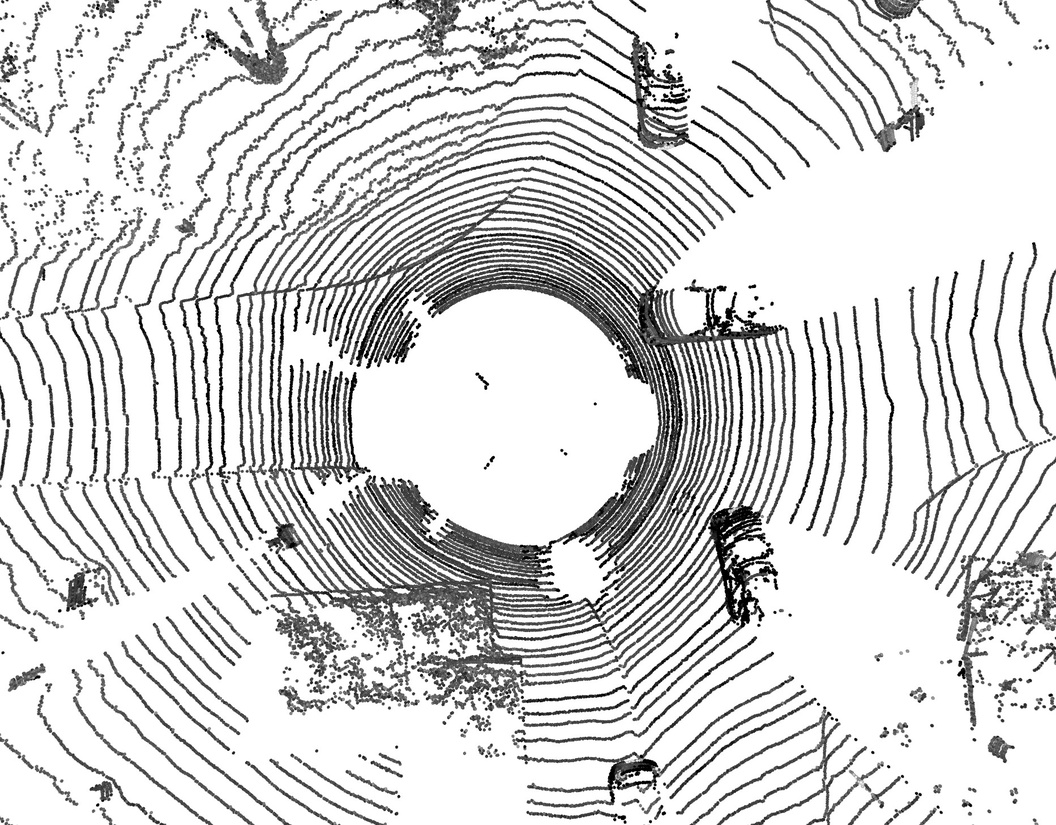}
    &
    \includegraphics[width=\linewidth]{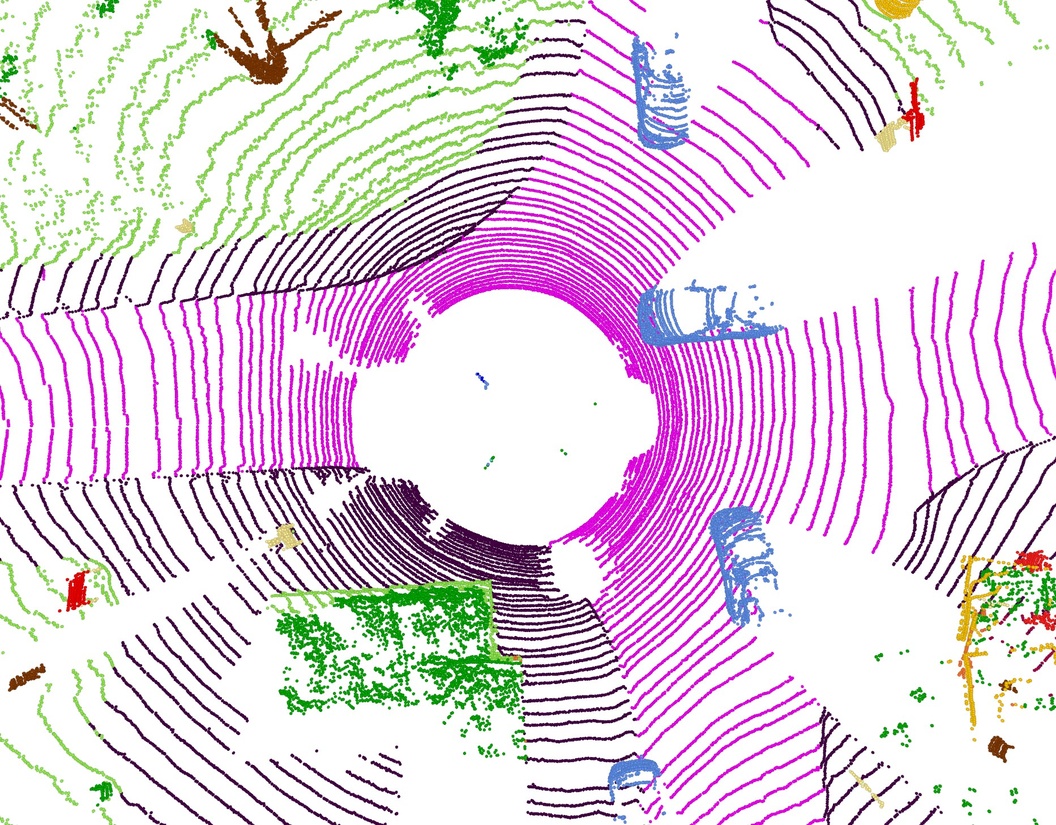}
    &
    \includegraphics[width=\linewidth]{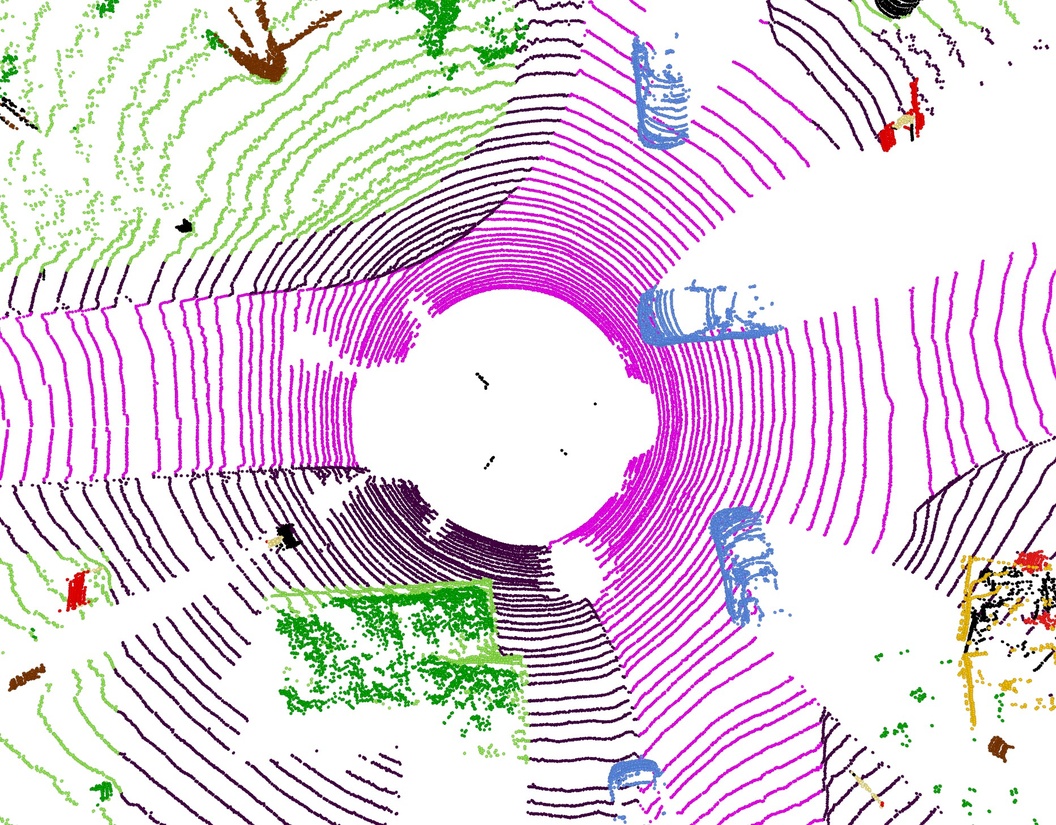}
    \\

    \includegraphics[width=\linewidth]{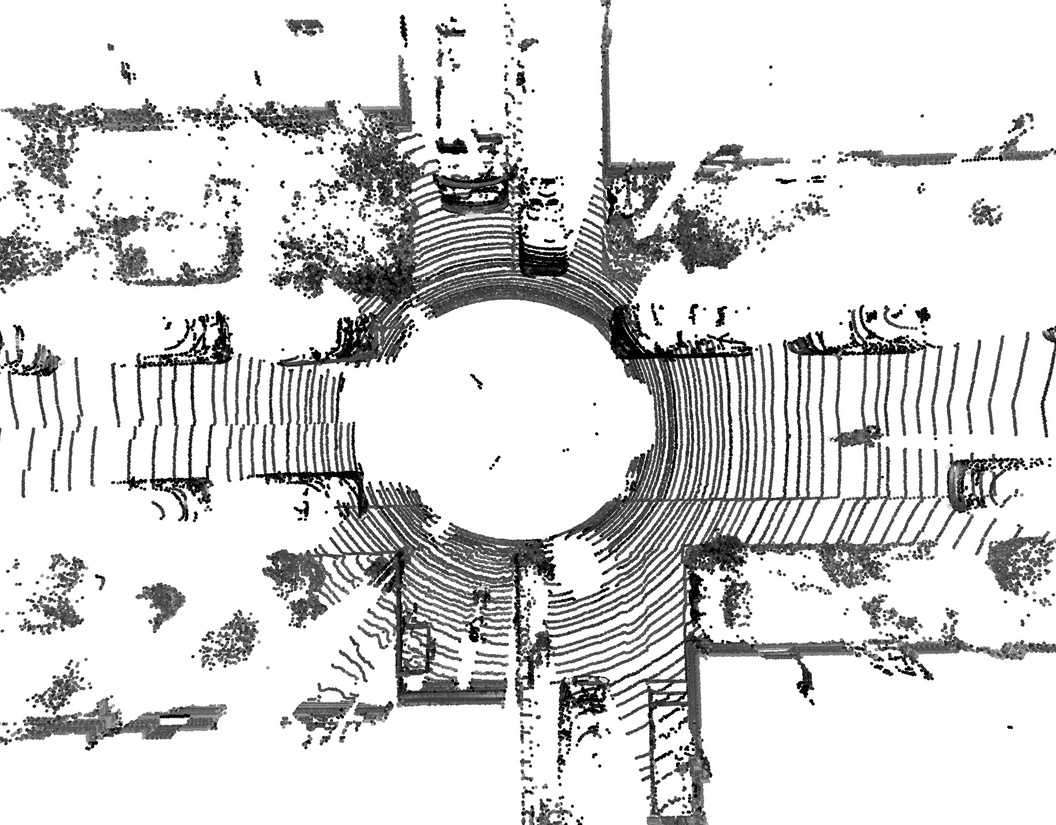}
    &
    \includegraphics[width=\linewidth]{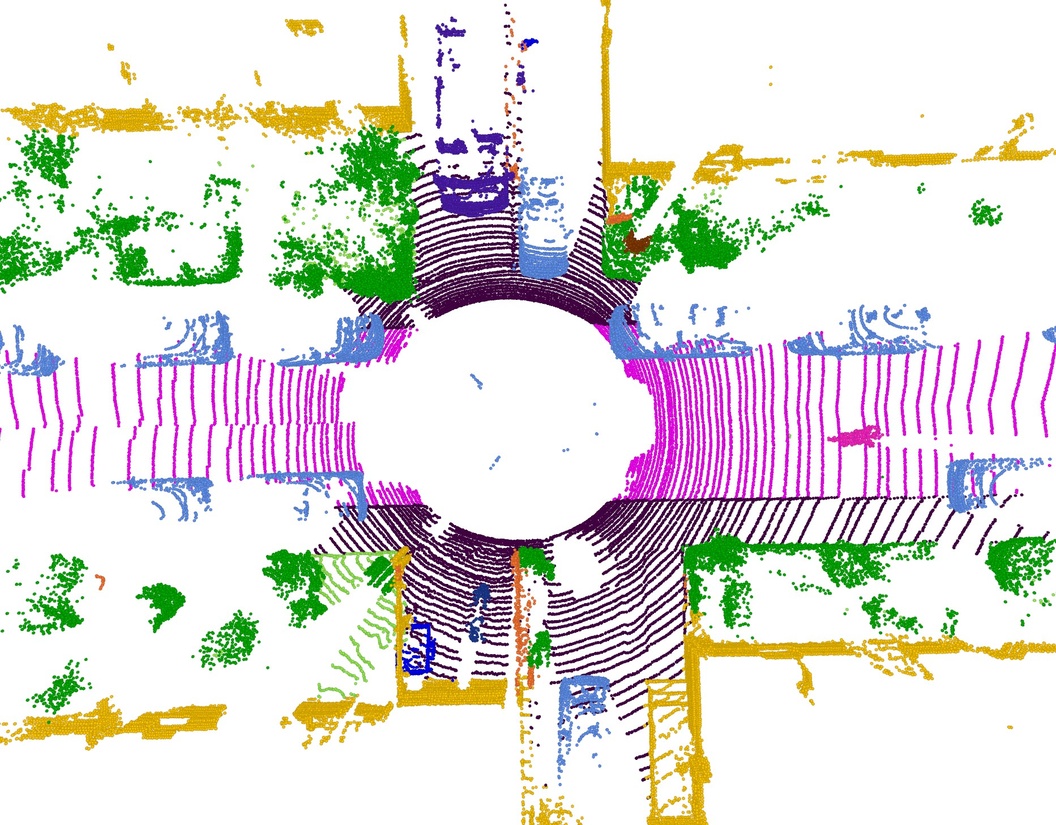}
    &
    \includegraphics[width=\linewidth]{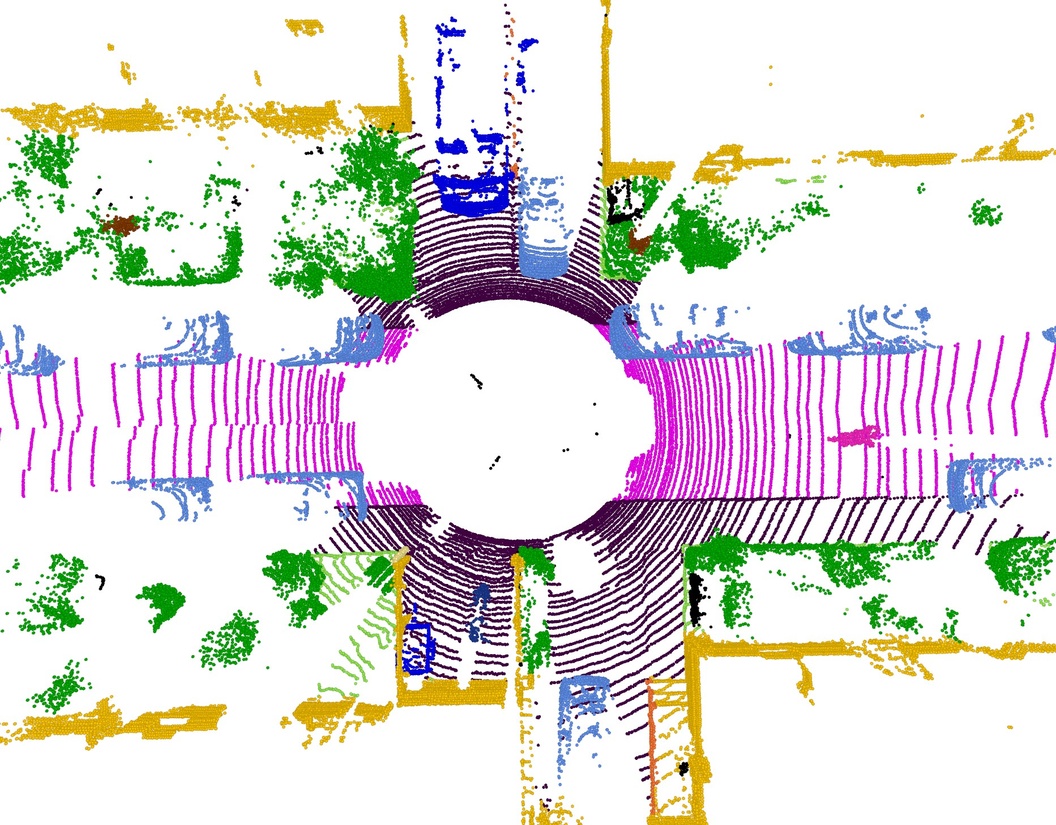}
    \\

    \includegraphics[width=\linewidth]{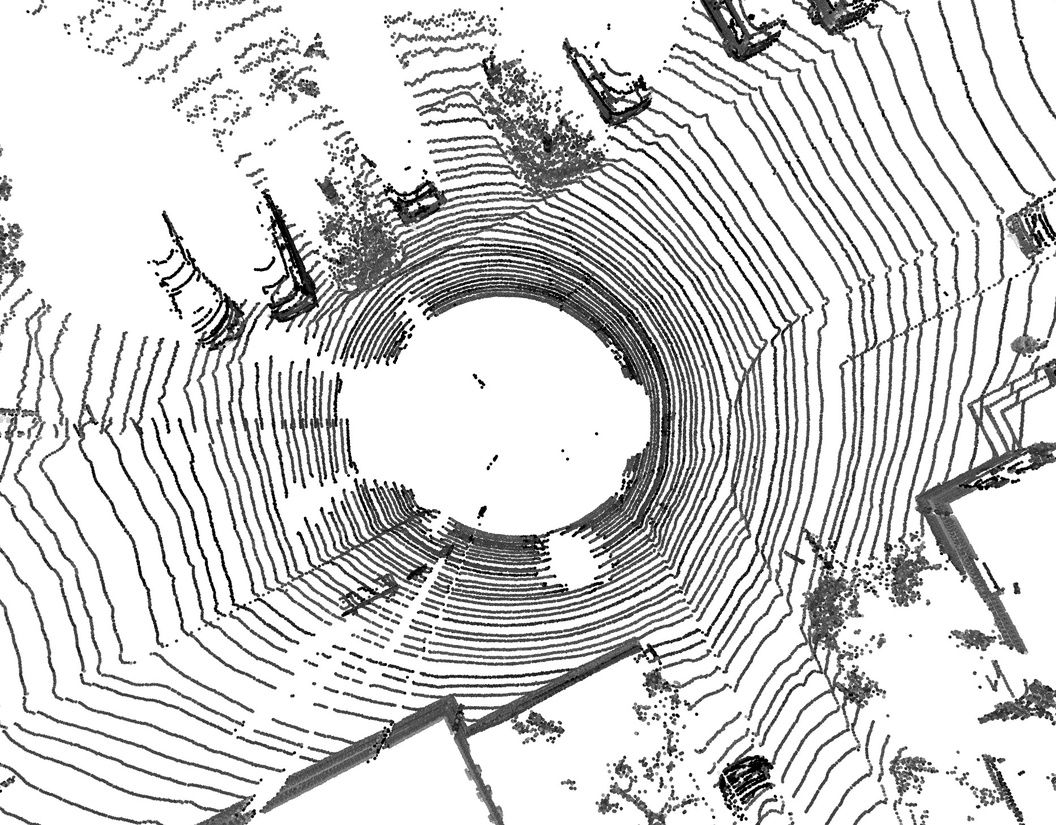}
    &
    \includegraphics[width=\linewidth]{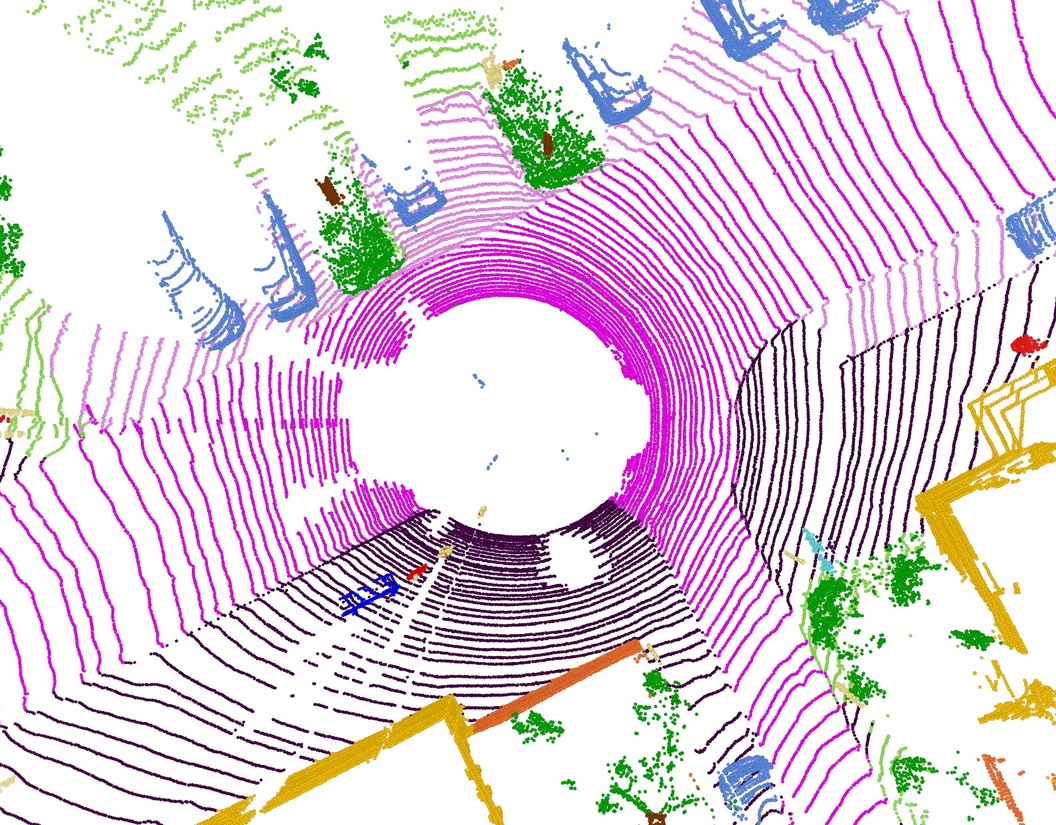}
    &
    \includegraphics[width=\linewidth]{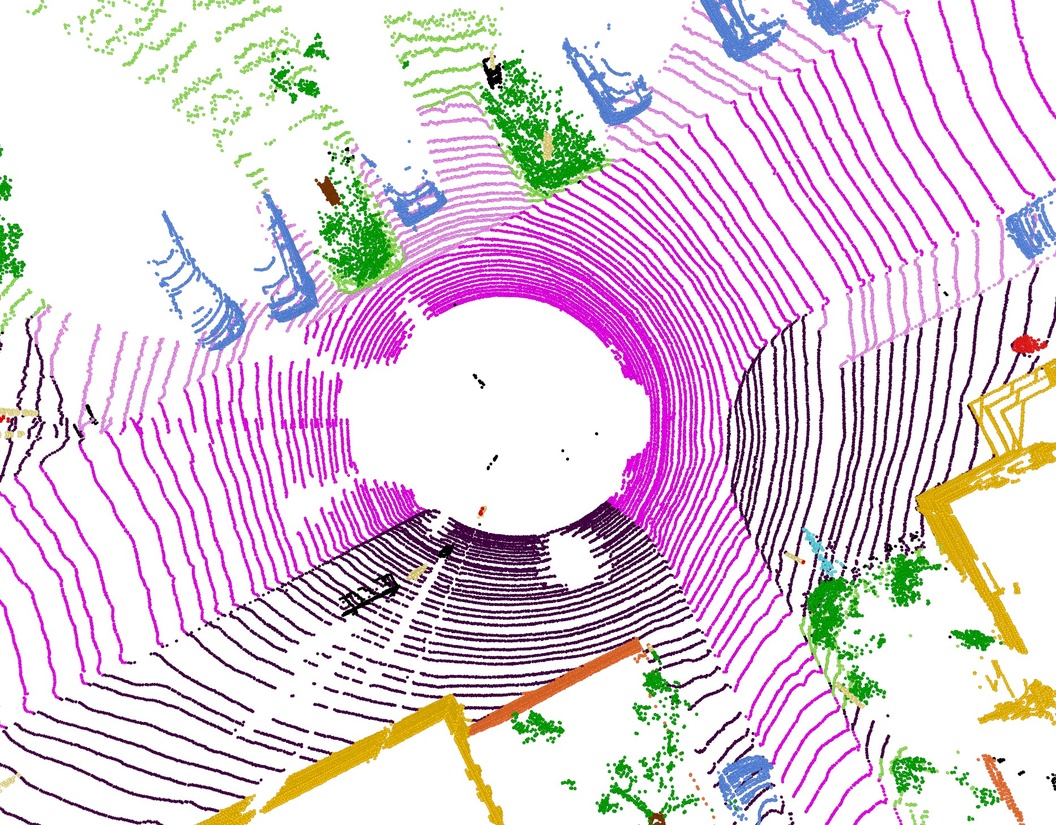}
    \\

    \includegraphics[width=\linewidth]{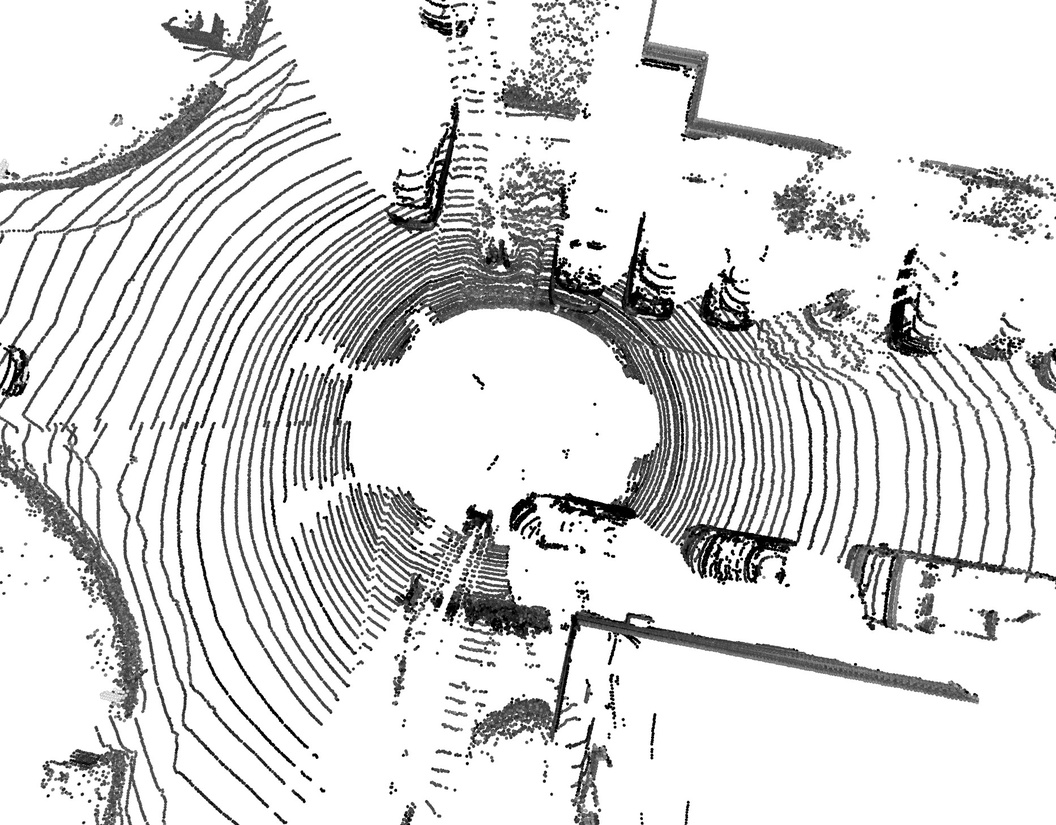}
    &
    \includegraphics[width=\linewidth]{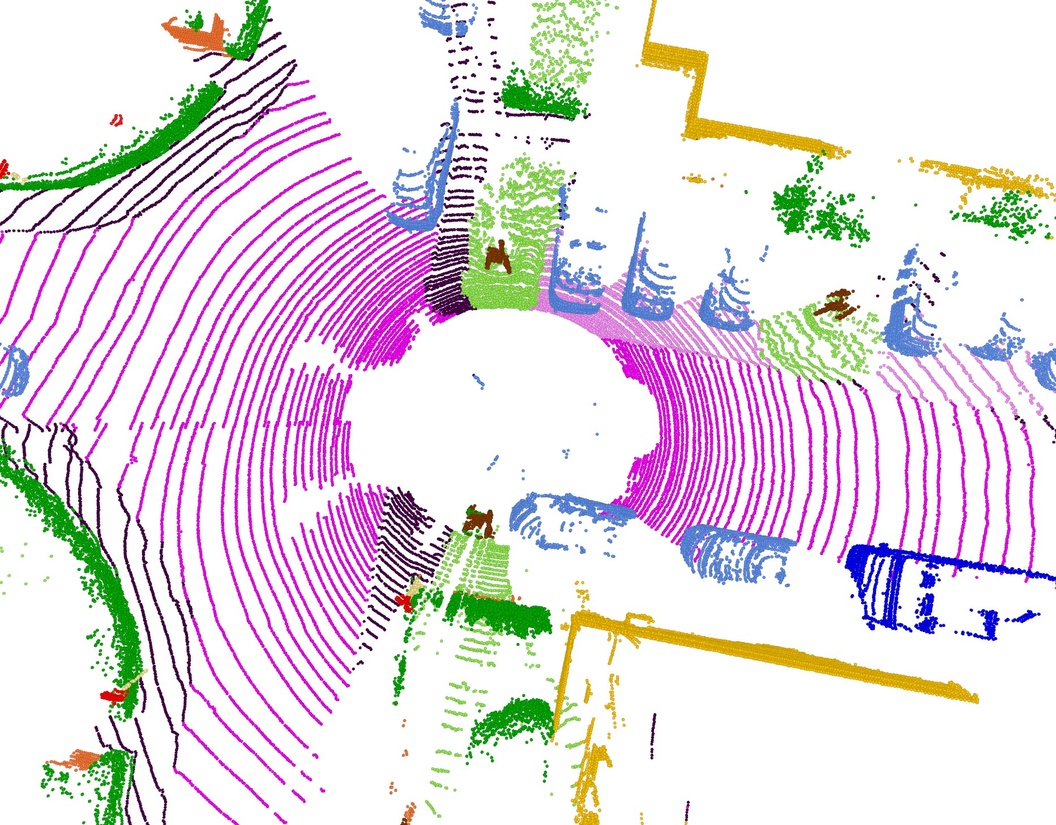}
    &
    \includegraphics[width=\linewidth]{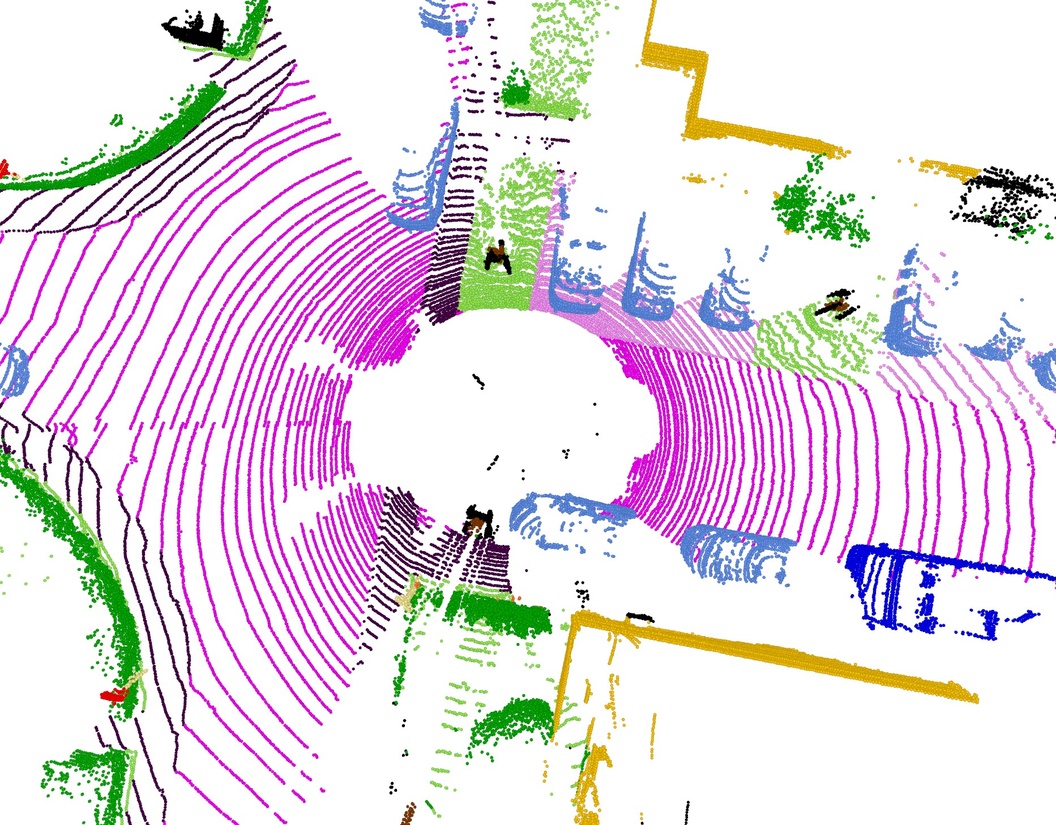}
    \\

    \includegraphics[width=\linewidth]{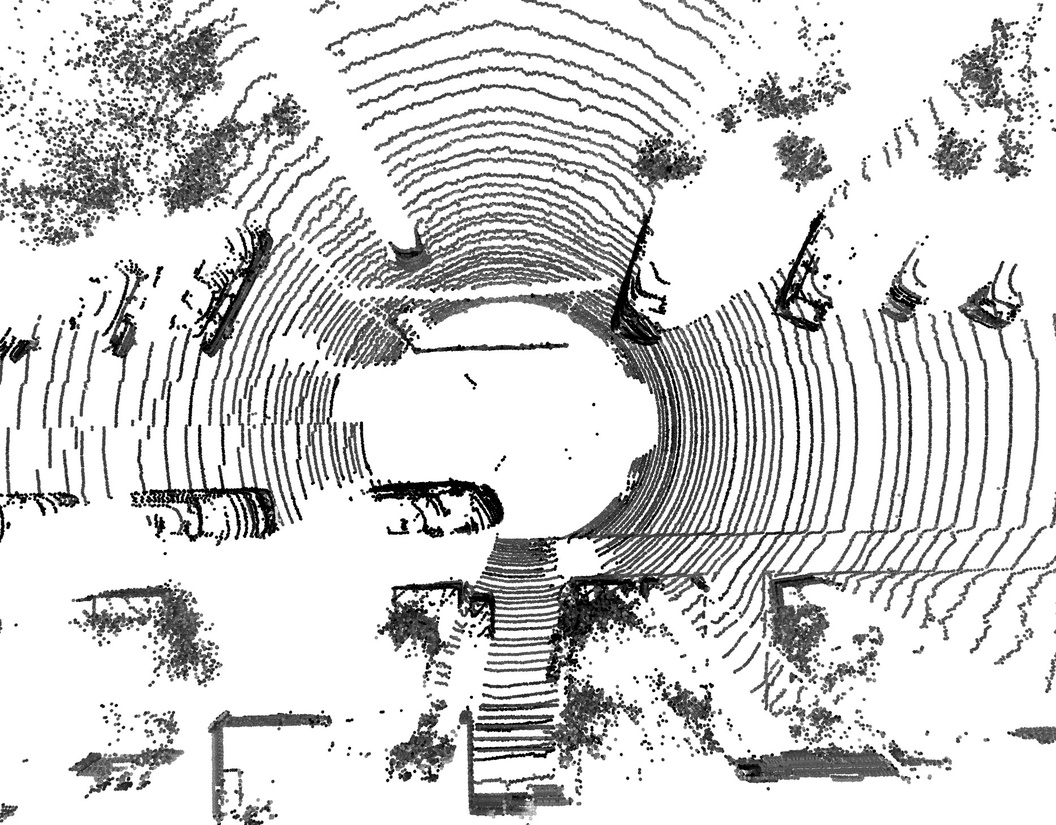}
    &
    \includegraphics[width=\linewidth]{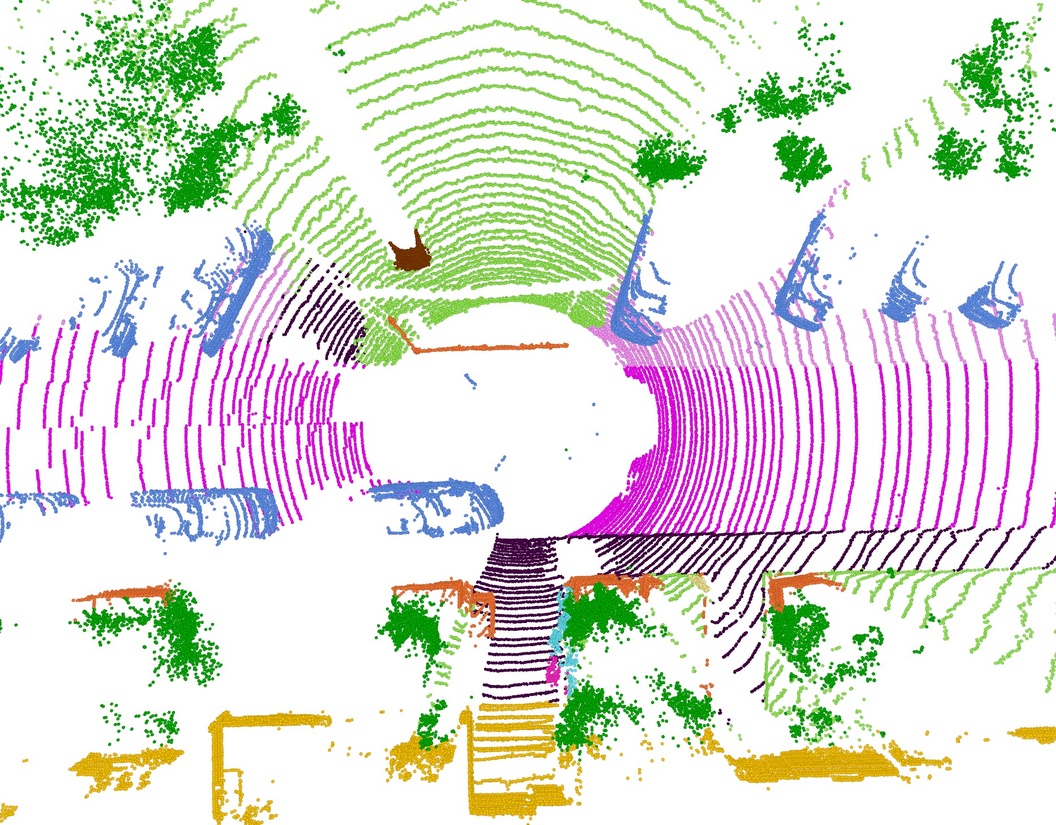}
    &
    \includegraphics[width=\linewidth]{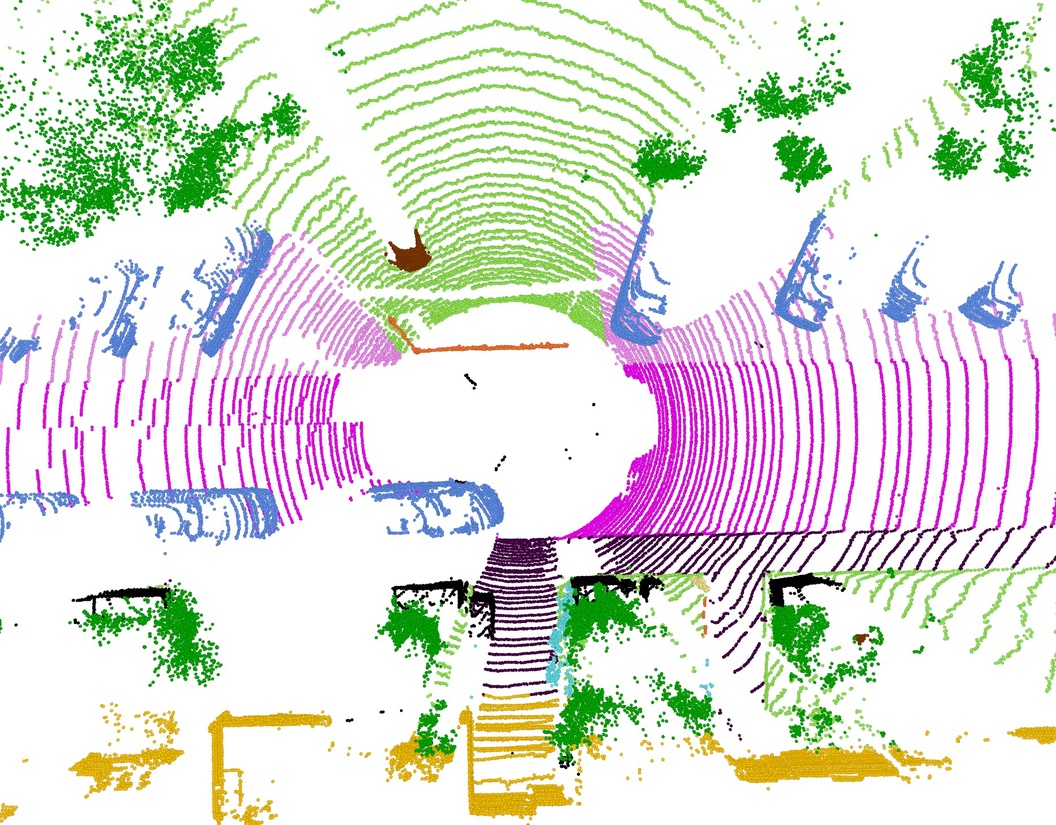}
    \\

\\[2ex]
    \multicolumn{3}{c}{\begin{minipage}{\textwidth}
    \centering
    \colorsquare{kitti_car}{kitti_car}\,car \quad
    \colorsquare{kitti_bicycle}{kitti_bicycle}\,bicycle \quad
    \colorsquare{kitti_motorcycle}{kitti_motorcycle}\,motorcycle \quad
    \colorsquare{kitti_truck}{kitti_truck}\,truck \quad
    \colorsquare{kitti_othervehicle}{kitti_othervehicle}\,other-vehicle \quad
    \colorsquare{kitti_person}{kitti_person}\,person \quad
    \colorsquare{kitti_bicyclist}{kitti_bicyclist}\,bicyclist \quad
    \colorsquare{kitti_motorcyclist}{kitti_motorcyclist}\,motorcyclist \quad
    \colorsquare{kitti_road}{kitti_road}\,road \quad
    \colorsquare{kitti_parking}{kitti_parking}\,parking \quad
    \colorsquare{kitti_sidewalk}{kitti_sidewalk}\,sidewalk \quad
    \colorsquare{kitti_otherground}{kitti_otherground}\,other-ground \quad
    \colorsquare{kitti_building}{kitti_building}\,building \quad
    \colorsquare{kitti_fence}{kitti_fence}\,fence \quad
    \colorsquare{kitti_vegetation}{kitti_vegetation}\,vegetation \quad
    \colorsquare{kitti_trunk}{kitti_trunk}\,trunk \quad
    \colorsquare{kitti_terrain}{kitti_terrain}\,terrain \quad
    \colorsquare{kitti_pole}{kitti_pole}\,pole \quad
    \colorsquare{kitti_trafficsign}{kitti_trafficsign}\,traffic-sign \quad
    \colorsquare{kitti_ignore}{kitti_ignore}\,ignore
    \end{minipage}}

    \end{tabularx}
    \caption{\textbf{Qualitative semantic segmentation results on SemanticKITTI.}}
    \label{fig:qualitative_kitti}
\end{figure*}

We present qualitative semantic segmentation results for the ScanNet and SemanticKITTI datasets in Fig.~\ref{fig:qualitative_scannet} and Fig.~\ref{fig:qualitative_kitti}, respectively.
Each example shows the input point cloud together with our predicted semantic labels and the corresponding ground-truth segmentation.
For these visualizations, we use \ours{}-B checkpoints obtained from multi-dataset training.
The results provide a qualitative comparison between predictions and annotations and further illustrate the quality of \ours{}'s predictions.
\section{Conclusion}
We revisit the vanilla Transformer encoder as a backbone for 3D scene understanding and show that, with few targeted adaptations, it can serve as a simple, scalable alternative to current specialized 3D architectures.
A key challenge is that Transformers generalize poorly under the limited supervision of current 3D benchmarks.
We address this with a data-efficient training recipe and show that this architecture benefits more from increased supervision, achieving new best results across 3D benchmarks while remaining more efficient than prior state-of-the-art methods.
Looking ahead, Volt’s architectural simplicity and strong scaling behavior suggest a promising path toward further unification of 3D scene understanding with the broader Transformer ecosystem.
We expect this work to unlock new possibilities for larger-scale 3D supervision and pretraining, seamless multimodal extensions with other Transformers, and continued progress driven by rapidly improving Transformer software and hardware support.

{
\small
\PAR{Acknowledgements.}
We acknowledge funding by BMFTR project ``WestAI'' (grant no.~16IS22094D) and the EU project ``JUPITER AI Factory'' (grant no.~101250682).
Computations were performed using resources granted by RWTH Aachen under projects \texttt{rwth1742}, \texttt{rwth1788}, and \texttt{rwth1968} and by the Gauss Centre for Supercomputing e.V. through the John von Neumann Institute for Computing on the GCS Supercomputer JUWELS at the J\"ulich Supercomputing Centre.
}
{
    \small
    \bibliographystyle{IEEEtranN}
    \bibliography{main}
}

\end{document}